\documentclass[sn-mathphys,Numbered]{sn-jnl}

\usepackage{graphicx}%
\usepackage{multirow}%
\usepackage{amsmath,amssymb,amsfonts}%
\usepackage{amsthm}%
\usepackage{mathrsfs}%
\usepackage[title]{appendix}%
\usepackage{textcomp}%
\usepackage{manyfoot}%
\usepackage{booktabs}%
\usepackage{algorithm}%
\usepackage{algorithmicx}%
\usepackage{algpseudocode}%
\usepackage{listings}%
\usepackage{subcaption}%
\usepackage{caption}%
\usepackage{tabularx}
\usepackage{float}
\usepackage{tikz}
\usepackage{pifont} 
\usepackage{lineno}
\usetikzlibrary{positioning, arrows.meta}
\usepackage{xcolor}
\usepackage[normalem]{ulem}
\usepackage{changes}
\setdeletedmarkup{\textcolor{red}{\sout{#1}}}
\usepackage{amssymb}
\usepackage{bbm}

\renewcommand{\textcolor}[2]{#2}
\renewcommand{\sout}[1]{\unskip}
\newcommand{\bcenter}[2]{\text{center}(\mathbf{b}_{#1}^{#2})} 

\usepackage[T1]{fontenc}

\begin{document}
\title[Article Title]{Mesh-Aware Epipolar Matching for Multi-View Multi-Person 3D Pose Estimation in Basketball}  

\author[1]{\fnm{Li} \sur{Yin}}\email{li.yin@g.sp.m.is.nagoya-u.ac.jp}
\author[1]{\fnm{Qin} \sur{Haobin}}\email{qin.haobin@g.sp.m.is.nagoya-u.ac.jp}
\author[1]{\fnm{Tomohiro} \sur{Suzuki}}\email{suzuki.tomohiro@g.sp.m.is.nagoya-u.ac.jp}
\author[1]{\fnm{Calvin} \sur{Yeung}}\email{yeung.chikwong@g.sp.m.is.nagoya-u.ac.jp}
\author[2]{\fnm{Mariko} \sur{Isogawa}}\email{mariko.isogawa@keio.jp}

\author*[1,3]{\fnm{Keisuke} \sur{Fujii}}\email{fujii@i.nagoya-u.ac.jp}

\affil[1]{\orgdiv{Graduate School of Informatics}, \orgname{Nagoya University}, \orgaddress{\street{Chikusa-ku}, \city{Nagoya}, \state{Aichi}, \country{Japan}}}

\affil[2]{\orgdiv{Department of Information Engineering, Faculty of Science and Technology}, \orgname{Keio University}, \orgaddress{\street{Hiyoshi}, \city{Yokohama}, \state{Kanagawa}, \country{Japan}}}

\affil[3]{\orgdiv{RIKEN Center for Advanced Intelligence Project}, \orgname{1-5}, \orgaddress{\street{Yamadaoka}, \city{Suita}, \state{Osaka},  \country{Japan}}}


\abstract
{Multi-view multi-person 3D pose estimation in team sports scenarios remains challenging due to players' occlusions, appearance similarity from team uniforms, and the scarcity of annotated multi-view data, which limits the effectiveness and generalization of learning-based methods. Conversely, the accuracy of training-free methods is inherently limited by the quality of 2D keypoints detection and the robustness of cross-view association. To address these challenges, we propose Mesh-Aware Epipolar Matching (MAEM), a training-free framework for multi-view multi-person 3D pose estimation. Our method leverages a monocular 3D human mesh recovery model as the frontend. Based on 3D human mesh recovery model outputs, we develop a two-stage epipolar matching strategy that combines disjoint set union-based clustering and per-joint triangulation, enabling robust cross-view association and accurate 3D pose reconstruction. Experiments on two public multi-view basketball datasets show that MAEM improves over training-free association baselines and achieves competitive RGB-only accuracy across indoor and outdoor basketball scenarios. MAEM achieves mean MPJPE/PA-MPJPE of 59.8/40.7 mm on SportCenter EPFL and 74.0/51.8 mm on Human-M3 Basketball, demonstrating the practical utility of dense mesh geometry for cross-view association without target-domain training or fine-tuning. The code and data are available at: \url{https://github.com/Yinlipp/MAEM.git}} 

\keywords{multi-person 3D pose estimation, multi-view fusion, epipolar geometry, body mesh recovery, team sports, training-free}

\maketitle
\section{Introduction}\label{Introduction}
3D human pose estimation has become an essential component of modern sports 
analytics~\cite{bridgeman2019multi, yeung2025athletepose3d, dong2019fast}, 
underpinning applications such as automated tactical analysis, biomechanical 
monitoring, and broadcast production. Team sports such as basketball and football represent one of the most demanding instantiations of this problem: 
high player density, frequent mutual occlusions, near-identical team uniforms, and rapid unpredictable motion combine to make multi-view multi-person 3D pose 
estimation substantially more challenging than in controlled 
benchmarks~\cite{Joo_2015_ICCV, belagiannis20143d}. Top-tier leagues address occlusion by deploying dozens of cameras throughout their venues, but such infrastructure is unavailable in most real-world settings. For the technology to generalize beyond elite competitions, robust multi-view multi-person 3D pose estimation must be achievable with only a sparse camera setup, placing the burden squarely on the cross-view association algorithm rather than on camera density.

The central challenge in multi-view multi-person 3D pose estimation is cross-view association: determining which detections across views correspond to the same physical person. In team sports, this problem is uniquely acute, as occlusion, large camera-to-subject distances, and near-identical uniforms cause conventional association cues to degrade simultaneously. Explicit association methods~\cite{zhang20204d, dong2019fast, bridgeman2019multi} 
establish cross-view correspondences using three types of cues: sparse skeletal 
keypoints, appearance-based re-identification~\cite{he2023fastreid, 
he2021transreid}, and epipolar geometric constraints. In team sports, however, all three cues fail simultaneously: keypoints are frequently occluded by other players, appearance features become indistinguishable under identical team jerseys, and epipolar constraints lose discriminative power when cameras are positioned far from the playing field or occluded. Learning-based methods~\cite{tu2020voxelpose, zhang2021direct, liao2024multiple, qiu2023psvt, huang2020end, lin2021multi, ye2022faster} bypass explicit cross-view matching by aggregating multi-view image features within a unified 3D representation space, jointly resolving person detection and cross-view association in a single forward pass. However, they require annotated training data and are sensitive to changes in camera configuration, both of which are practical barriers in team sports where labeled multi-view data are scarce~\cite{sportcenter_multiview, fan2023human, jiang2024worldpose}. Body mesh models, which encode dense geometric structure of the human body, offer a promising alternative cue that is invariant to appearance, yet their potential for cross-view association remains largely untapped. Recent attempts to introduce body mesh priors into this pipeline \cite{dong2021shape, hokari2025human} have shown promise in correcting implausible poses and improving cross-view consistency. Nevertheless, existing mesh-aware approaches either invoke the body model only as a post-hoc refinement after association decisions have already been made, or remain unvalidated under the densely crowded, uniform-appearance conditions of competitive team sports. Consequently, robust multi-view multi-person 3D pose estimation in team sports remains an open challenge, as explicit methods break down under uniform appearance and occlusion, while learning-based methods are hindered by scarce annotations and sensitivity to camera configurations.

We propose MAEM (Mesh-Aware Epipolar Matching), a training-free framework for multi-view multi-person 3D pose estimation in team sports, to address the challenge of cross-view person matching. The core idea is to use projected dense body mesh vertices as epipolar verification cues for cross-view association, departing from conventional explicit methods that rely on sparse skeletal keypoints or appearance features, both of which fail under occlusion and uniform team jerseys. Dense mesh vertices provide substantially richer geometric coverage of the human body surface, making epipolar correspondences robust precisely where keypoint-based approaches break down. Although body mesh models have been applied to pose refinement and mesh fusion in prior work, their direct use as a cross-view association primitive in a training-free setting has not been explored, nor validated under the densely crowded, uniform-appearance conditions of competitive team sports. The main contributions of this work are as follows:

\begin{itemize}
    \item[$\bullet$] A pipeline that directly exploits projected dense monocular mesh vertices as epipolar verification cues within the cross-view association stage, enabling robust multi-person matching and generalizing across team sports scenarios.

    \item[$\bullet$] A two-stage matching mechanism combining a coarse bounding-box center reprojection filter and a mesh-vertex epipolar distance filter, which together suppress false correspondences more effectively than sparse-keypoint alternatives, as confirmed by ablation studies.

    \item[$\bullet$]MAEM achieves the lowest MPJPE of 74.0~mm on the Human-M3 basketball scenarios benchmark~\cite{fan2023human} and 59.8~mm on the Sportcenter dataset~\cite{sportcenter_multiview} under uniform-appearance conditions, with MPJPE reductions of 19.1\% and 9.0\% over the top-performing training-free association baseline, respectively, demonstrating consistent generalization across indoor and outdoor team sports scenarios under varying camera configurations, without any target-domain network training. 
\end{itemize}

\section{Related work}\label{Related work} 

\subsection{Explicit association multi-view 3D pose estimation}
Early approaches to multi-view multi-person 3D pose estimation~\cite{belagiannis20143d, bridgeman2019multi, zhang20204d, dong2019fast} adopt a sequential bottom-up pipeline that explicitly matches detected 2D observations across camera views before reconstructing 3D poses. Belagiannis et al.~\cite{belagiannis20143d} introduce a 3D Pictorial Structures model that represents the human body as a set of rigid parts connected by pairwise potentials, enabling joint inference of part locations across multiple views within a unified probabilistic framework. By reasoning over individual body parts rather than holistic detections, this approach tolerates partial occlusion more gracefully than whole-body matching methods. Dong et al.~\cite{dong2019fast} propose a multi-way matching algorithm to cluster detected 2D poses across all views by combining appearance-based re-identification features~\cite{he2023fastreid, he2021transreid} and epipolar geometric constraints, from which the 3D pose can be effectively inferred through triangulation. Whereas this approach is fast and robust against missing and false detections, it remains fundamentally sensitive to the quality of upstream 2D pose detections, and appearance-based matching degrades significantly when individuals share similar visual attributes. Furthermore, epipolar constraints derived from sparse skeletal keypoints lose discriminative power under heavy occlusion, as partially visible skeletons yield ambiguous geometric evidence.

Zhang et al.~\cite{zhang20204d} unify per-view parsing, cross-view matching, and temporal tracking into a single 4D association graph optimization framework, treating each dimension of image space, viewpoint, and time equally and simultaneously. Although this formulation enables more coherent multi-person tracking across frames, the joint optimization of four dimensions incurs substantial computational complexity, limiting its scalability to large-scale sports environments. Zhou et al.~\cite{zhou2022quickpose} formulate the multi-view matching problem as mode seeking in the space of skeleton proposals, where each skeleton may consist of an arbitrary number of body parts rather than requiring full-body or single-part representations, enabling real-time motion capture in crowded scenes. Yet, mode seeking implicitly assumes that correct correspondences form distinguishable density peaks in the proposal space, an assumption that breaks down when multiple players adopt similar poses or when severe occlusion reduces the number of available joint candidates, both of which are common occurrences in competitive team sports. Bridgeman et al.~\cite{bridgeman2019multi} further demonstrate that combining geometric and appearance cues improves association robustness in sports scenarios, yet their method still relies on discriminative appearance features that become unreliable under identical team uniforms.

Despite their differences in formulation, all these methods share a common vulnerability: their association cues, whether appearance, sparse keypoints, or density peaks, degrade simultaneously under the uniform-appearance and heavy-occlusion conditions characteristic of competitive team sports. MAEM directly addresses this limitation by substituting these fragile signals with projected dense mesh vertices, which provide richer and more occlusion-robust geometric evidence for cross-view association.

\subsection{Learning-based, self-supervised, and generalizable multi-view 3D pose estimation}

To eliminate the need for hand-crafted cross-view matching heuristics, a parallel line of work develops end-to-end learning frameworks that jointly resolve person detection and multi-view association within a trainable network ~\cite{tu2020voxelpose, zhang2021direct, liao2024multiple, qiu2023psvt, huang2020end, lin2021multi, ye2022faster}. Tu et al.~\cite{tu2020voxelpose} aggregate projected 2D heatmaps from all camera views into a unified 3D voxel space, which is then processed by a Cuboid Proposal Network and a Pose Regression Network to localize and refine 3D poses without explicit cross-view association. Zhang et al.~\cite{zhang2021direct} present the Multi-view Pose Transformer (MvP), which represents skeleton joints as learnable query embeddings that progressively attend to multi-view image features via a projective attention mechanism, directly regressing 3D joint locations without relying on any intermediate matching task. Huang et al.~\cite{huang2020end} introduce an end-to-end dynamic matching network that jointly optimizes cross-view association and 3D pose estimation in a differentiable manner. Although these learning-based methods achieve strong performance on controlled indoor benchmarks, they are highly dependent on large-scale multi-view annotated training data and remain sensitive to changes in camera configuration, limiting their applicability to diverse sporting venues where labeled data are scarce.

To reduce reliance on 3D ground-truth annotations, self-supervised approaches have emerged as a promising alternative. Wandt et al.~\cite{wandt2021canonpose} propose CanonPose, a self-supervised framework that learns a monocular 3D pose estimator from unlabeled multi-view data by exploiting multi-view consistency constraints to disentangle the observed 2D pose into the underlying 3D pose and camera rotation, notably without requiring calibrated cameras. Srivastav et al.~\cite{srivastav2024selfpose3d} further advance this direction by proposing a self-supervised multi-person framework that leverages multi-view geometric consistency as a training signal, achieving competitive performance without requiring expensive 3D supervision. While these approaches substantially lower the annotation burden, they still require scene-specific training and remain sensitive to significant shifts in camera configuration, limiting their out-of-the-box applicability to new sporting venues where neither labeled data nor stable camera setups are guaranteed.

Generalizing across diverse camera configurations remains a fundamental challenge for learning-based multi-view pose estimation. Bartol et al.~\cite{bartol2022generalizable} propose a stochastic triangulation framework for human pose estimation that explicitly addresses generalizability across different camera arrangements and numbers of cameras, achieving substantial improvements over existing methods on novel camera configurations. Ye et al.~\cite{ye2022faster} further improve generalization by decoupling person localization and pose estimation in a volumetric framework, reducing computational overhead while maintaining competitive accuracy across different camera setups. Liao et al.~\cite{liao2024multiple} introduce a graph-based association mechanism that adapts to varying numbers of cameras without retraining. Nevertheless, even these more flexible methods require domain-specific supervision during training and assume access to calibrated multi-view data from similar environments, falling short of true training-free generalization. In contrast, MAEM requires neither training data nor target-domain network training, achieving robust cross-view association across diverse team sports scenarios through mesh-vertex epipolar matching. 

\subsection{Body Mesh Priors for Multi-View Pose Estimation}
Several works have explored parametric body mesh models as a means to introduce 3D structural priors into multi-view pose estimation. Dong et al.~\cite{dong2021shape} apply SMPL~\cite{SMPL:2015} as a post-hoc refinement after triangulation to correct implausible poses and fill missing joints, yet the iterative optimization is computationally expensive and cannot resolve association ambiguities at the matching stage itself. Hokari et al.~\cite{hokari2025human} take a different approach by reconstructing human meshes independently from each dynamic camera view and fusing them into a global coordinate system via confidence-based weighting. This calibration-free framework demonstrates promising results in everyday sports scenarios; however, its evaluation is confined to small-scale, low-density settings and has not been validated under the densely crowded, uniform-appearance conditions of competitive team sports. Unlike these approaches, MAEM embeds mesh-derived geometric constraints directly within the matching stage, enabling appearance-agnostic cross-view association without iterative optimization.

More recent approaches extend mesh priors to uncalibrated setups and close interactions. Yin et al.~\cite{yin2025easyret3d} leverage pre-trained mesh recovery models and ground plane constraints for camera extrinsic estimation without manual calibration, though their focus is on camera parameter recovery rather than cross-view association. Lu et al.~\cite{lu2024avatarpose} use personalized neural avatars to guide pose estimation of closely interacting individuals, but per-subject avatar optimization limits scalability to large player groups in team sports.

Unlike these methods, MAEM embeds dense mesh vertices directly as epipolar verification cues within the association stage, addressing cross-view matching at its source rather than as a post-hoc correction, without per-subject optimization or iterative refinement.
  
\subsection{Monocular 3D Human Mesh Recovery as a Frontend}

Monocular human mesh recovery methods reconstruct the full 3D surface geometry of the human body from a single RGB image using parametric models such as SMPL~\cite{SMPL:2015} and SMPL-X~\cite{pavlakos2019expressive}. By encoding the body in a low-dimensional parameter space, these models embed strong anatomical priors that enforce kinematic plausibility and enable the recovery of occluded joints by leveraging global body-configuration constraints. SAM 3D Body~\cite{yang2026sam3dbody} represents the current state of the art, introducing the Momentum Human Rig \cite{MHR:2025} representation that decouples skeletal structure from surface shape, and demonstrating superior robustness to occlusions and challenging viewpoints through a promptable encoder-decoder architecture. 

In MAEM, monocular mesh recovery serves solely as a front-end perception module: the dense surface vertices predicted per-view by SAM 3D Body are treated as geometric anchors and passed directly to the cross-view association stage. The primary contribution of this work lies not in mesh reconstruction itself, but in how the recovered mesh geometry is exploited to achieve robust appearance-agnostic cross-view association in competitive team sports.

\subsection{Sports Pose Estimation Datasets and Benchmarks}

The development of sports pose estimation has been supported by a growing number of single-person and monocular datasets. SportsPose~\cite{ingwersen2023sportspose} captures over 176,000 3D poses from 24 subjects across five dynamic sports activities. AthletePose3D~\cite{yeung2025athletepose3d} and AthleticsPose~\cite{suzuki2025athleticspose} further extend coverage to competitive athletic movements in real outdoor environments. AutoSoccerPose~\cite{yeung2024autosoccerpose} provides 2D and 3D pose annotations extracted from professional soccer broadcast videos. While these datasets have advanced single-person sports pose estimation, they do not address the multi-person association and multi-view synchronization challenges inherent in team sports.

To bridge this gap, several datasets targeting multi-person team sports scenarios have been proposed. WorldPose~\cite{jiang2024worldpose} provides global 3D human pose annotations captured during the FIFA World Cup in a real outdoor football environment. TrackID3x3~\cite{yamada2025trackid3x3, yin2024enhanced} is the first publicly available dataset for multi-player tracking and 2D pose estimation in 3x3 basketball, covering indoor, outdoor, and drone camera perspectives. 

Despite this growing body of work,  the scarcity of 3D-annotated multi-view pose data remains the central bottleneck for multi-person team sports pose estimation. Obtaining ground-truth 3D poses requires either marker-based motion capture, which is impractical in real game conditions, or manual multi-view triangulation, which is prohibitively costly at scale. The Sportcenter dataset~\cite{sportcenter_multiview} and Human-M3~\cite{fan2023human} represent the two primary benchmarks for multi-view multi-person 3D pose estimation in basketball, yet both are limited in scale and camera diversity. This scarcity of multi-view basketball 3D pose data highlights the need for methods that can generalize without target-domain training or finetuning, which is precisely the gap that MAEM aims to address.

\section{Method}\label{Methods}
\begin{figure*}[t]
\centering
\includegraphics[width=\textwidth]{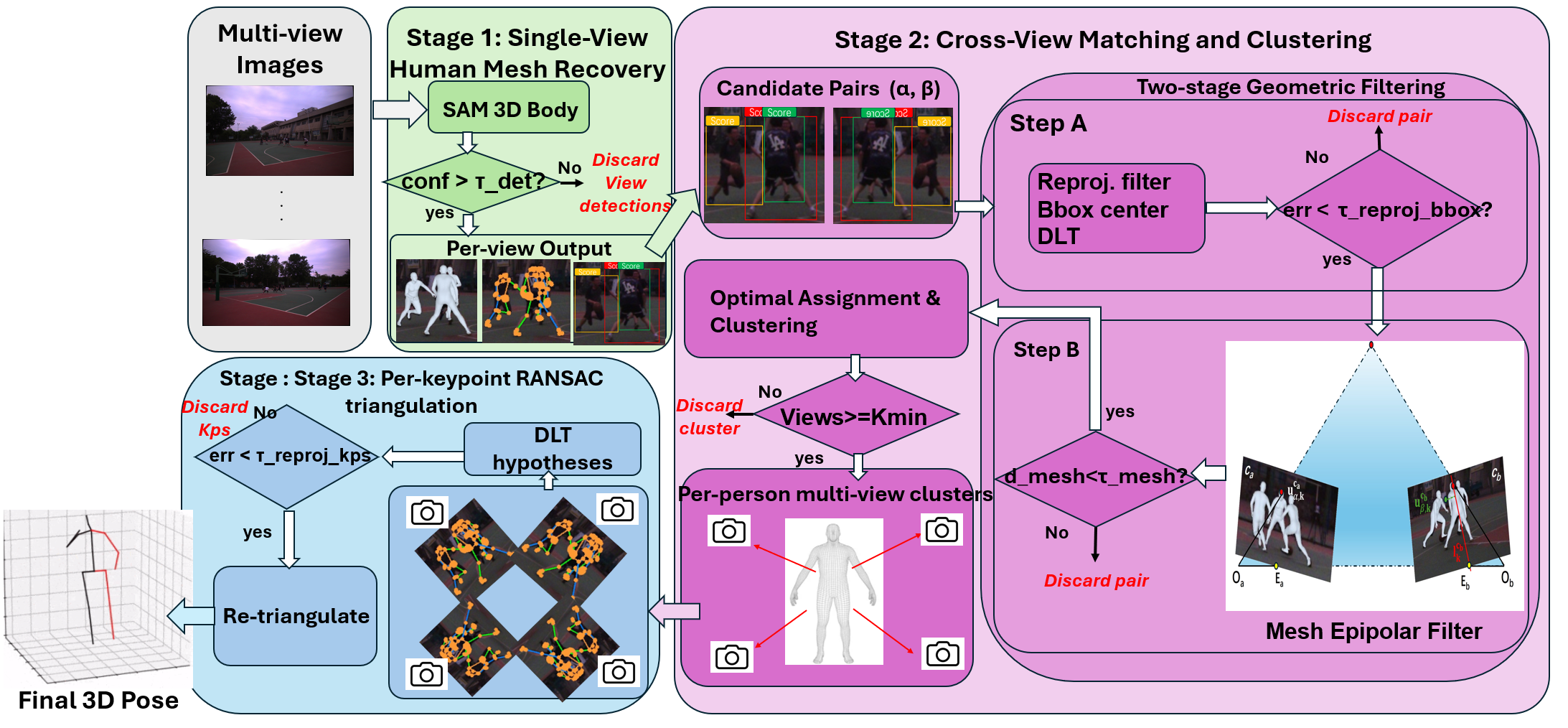}
\caption{Overview of the MAEM pipeline. Given multi-view images, Stage 1 recovers per-person 3D meshes and keypoints via SAM 3D Body. Stage 2 associates detections across views using a two-stage epipolar filter and Hungarian matching. Stage 3 reconstructs the final 3D pose via RANSAC triangulation.}
\label{fig:pipeline}
\end{figure*}

To address the cross-view association challenges inherent to team sports scenarios, we propose Mesh-Aware Epipolar Matching (MAEM), a training-free framework for robust multi-view multi-person 3D pose estimation under a calibrated multi-camera setup, as illustrated in Figure~\ref{fig:pipeline}. Supplementary Figure S1 provides a detailed flowchart of the MAEM pipeline.

In Stage~1 (Section~\ref{sec:single_view}), SAM 3D Body~\cite{yang2026sam3dbody} 
is applied independently to each camera view to predict per-person bounding boxes, 3D poses, and 3D body meshes, where the 3D poses and meshes are projected onto the image plane to obtain the corresponding 2D poses and 2D vertex projections. Where necessary, fisheye lens distortion is corrected to ensure geometric consistency. These outputs serve as the foundation for subsequent cross-view matching and triangulation.

In Stage~2 (Section~\ref{sec:matching}), detections across views are associated into per-person clusters. A lightweight reprojection filter first eliminates a large proportion of geometrically implausible candidate pairs, after which a dense mesh epipolar filter evaluates consistency over the projected 2D surface vertices to reject remaining ambiguous matches. Pairs selected by both filters are assigned via Hungarian matching~\cite{kuhn1955hungarian}, then merged into multi-view clusters through Union-Find, a disjoint-set data structure that efficiently groups related observations across views. Conflicting observations, defined as multiple detections from the same camera assigned to the same cluster, are removed based on the matching cost.

In Stage~3 (Section~\ref{sec:triangulation}), based on the person clustering results, cross-view keypoint correspondences are computed via RANSAC triangulation d to synthesize the final 3D pose, weighted by person detection confidence scores.

The entire pipeline is training-free and directly applicable to unseen sports venues.

\subsection{Single-View Human Mesh Recovery} \label{sec:single_view}

For each camera view $c$, MAEM applies SAM 3D Body~\cite{yang2026sam3dbody} to the input image. When fisheye lenses are used, images are undistorted using the known intrinsic and distortion parameters prior to inference, ensuring consistency with the pinhole camera model assumed by calibration. SAM 3D Body reconstructs a full body mesh with 18,439 vertices based on the Momentum Human Rig (MHR) parametric mesh representation~\cite{MHR:2025} for each detected person $i$, producing a rich set of outputs including bounding boxes, shape and pose parameters, 3D mesh vertices, and 3D skeletal joints in camera-relative coordinates, along with their 2D projections onto the image plane. Among these, MAEM utilizes four outputs in downstream stages: (1)~a bounding box $\mathbf{b}_i^c$ with detection confidence $s_i^c$, which serves three roles: detections with confidence below a threshold $\tau_{\text{det}}$ are discarded, the bounding box center is used as a representative point for the reprojection filter, and the confidence score is used as a per-view weight during triangulation; (2)~projected 2D keypoints $\{\mathbf{p}_{i,j}^c\}_{j=1}^{J}$, where $j \in \{1, \dots, J\}$ indexes the $J$ body keypoints, used for multi-view triangulation; (3)~3D keypoints $\{\mathbf{q}_{i,j}^c\}_{j=1}^{J}$ in camera-relative coordinates, used for Procrustes alignment when computing PA-MPJPE; and (4)~projected 2D surface vertices $\mathcal{U}_i^c = \{\mathbf{u}_{i,k}^c\}_{k=1}^{18439}$, which provide dense geometric cues for the mesh epipolar filter. Note that not every person is necessarily visible in all $N$ views; for a given person cluster, $N_c \leq N$ denotes the number of views in which that person is successfully detected, and is used in Stage 3 for triangulation.

\subsection{Cross-View Matching and Clustering}\label{sec:matching}

The central challenge of multi-view multi-person pose estimation lies in cross-view association: determining which detections across different views correspond to the same physical person. MAEM addresses this through a two-stage geometric filtering scheme followed by optimal assignment and graph-based clustering.
 
\noindent\textbf{Candidate Pair Construction.}
For every pair of camera views $(c_a, c_b)$, all detection pairs $(\alpha, \beta)$ where $\alpha$ is from view $c_a$ and $\beta$ is from view $c_b$ are considered as candidate matches.
 
\noindent\textbf{Step~A: Reprojection Filter.}
Each candidate pair is evaluated by a lightweight geometric test. 
The bounding box centers $\bcenter{\alpha}{c_a}$ and 
$\bcenter{\beta}{c_b}$ of detections $\alpha$ and $\beta$ are 
triangulated via the Direct Linear Transform (DLT) using the 
pre-calibrated camera projection matrices of views $c_a$ and $c_b$, 
yielding a candidate 3D point $\tilde{\mathbf{X}}_{\alpha\beta} \in 
\mathbb{R}^3$. This 3D point is reprojected back into both views, and 
the mean reprojection error $e_{\text{reproj}}^{\text{bbox}}$ (in pixels) 
is computed as:

\begin{equation}
    e_{\text{reproj}}^{\text{bbox}}(\alpha, \beta) = \frac{1}{2} \left(
    \left\| \bcenter{\alpha}{c_a} - 
    \pi_{c_a}(\tilde{\mathbf{X}}_{\alpha\beta}) \right\|_2 +
    \left\| \bcenter{\beta}{c_b} - 
    \pi_{c_b}(\tilde{\mathbf{X}}_{\alpha\beta}) \right\|_2
    \right)
\end{equation}

\noindent where $\pi_c(\cdot)$ denotes the projection function of camera $c$, accounting for both perspective and fisheye camera models.

\noindent\textbf{Step~B: Dense Mesh Epipolar Filter.}
For filtered candidate pairs $(\alpha, \beta)$, MAEM performs a geometric consistency check using the projected 2D surface vertices rather than sparse skeletal keypoints. In team sports, the limited number of skeletal keypoints are frequently occluded by other players, making standard keypoint-based epipolar matching unreliable. By contrast, the dense 2D surface vertices, which cover the full body surface including torso contours and limb boundaries, provide a more discriminative and occlusion-robust geometric signal for cross-view verification. Since MHR maintains a fixed mesh topology across all reconstructions, the vertex index $k$ consistently corresponds to the same anatomical location, enabling direct vertex-wise epipolar comparison. 

The fundamental matrix $\mathbf{F}_{c_a c_b}$ between views $c_a$ and $c_b$ 
is derived from the pre-calibrated intrinsic matrices $\mathbf{K}_{c_a}$, 
$\mathbf{K}_{c_b}$ and the relative rotation $\mathbf{R}_{c_a c_b}$ and 
translation $\mathbf{t}_{c_a c_b}$ between the two views~\cite{hartley2003multiple}:

\begin{equation}
    \mathbf{F}_{c_a c_b} = \mathbf{K}^{-\top}_{c_b} 
    [\mathbf{t}_{c_a c_b}]_{\times} \mathbf{R}_{c_a c_b} 
    \mathbf{K}^{-1}_{c_a}
\end{equation}

\noindent where $[\cdot]_{\times}$ denotes the skew-symmetric matrix of a vector.

For each projected vertex $\mathbf{u}_{\alpha,k}^{c_a}$ of detection $\alpha$ 
in view $c_a$, the corresponding epipolar line in view $c_b$ is computed as:

\begin{equation}
    \mathbf{l}_k^{c_b} = \mathbf{F}_{c_a c_b} \, \tilde{\mathbf{u}}_{\alpha,k}^{c_a}
\end{equation}

\noindent where $\tilde{\mathbf{u}}_{\alpha,k}^{c_a}$ denotes the homogeneous 
coordinates of $\mathbf{u}_{\alpha,k}^{c_a}$. The epipolar line 
$\mathbf{l}_k^{c_b} = (a, b, c)^\top$ defines a line $ax + by + c = 0$ in 
view $c_b$, clipped to the image boundary to obtain its two endpoints. 
The point-to-epipolar-line distance for vertex $k$ is then:

\begin{equation}
    \mathrm{dist}\!\left(\mathbf{u}_{\beta,k}^{c_b},\, 
    \mathbf{l}_k^{c_b}\right) = 
    \frac{\left| a u_{\beta,k}^{x} + b u_{\beta,k}^{y} + c \right|}{\sqrt{a^2 + b^2}}
\end{equation}


\noindent where $(u_{\beta,k}^{x}, u_{\beta,k}^{y})$ are the 2D coordinates 
of the projected vertex $\mathbf{u}_{\beta,k}^{c_b}$. The overall geometric 
consistency between detections $\alpha$ and $\beta$ is measured by the mean 
distance over all vertices:

\begin{equation}
    d_{\text{mesh}}(\alpha, \beta) = \frac{1}{N_v} 
    \sum_{k=1}^{N_v}
    \mathrm{dist}\!\left(\mathbf{u}_{\beta,k}^{c_b},\; 
    \mathbf{l}_k^{c_b}\right)
\end{equation}

\noindent where $N_v$ denotes the number of mesh vertices used for epipolar verification. $N_v$ =18,439 in our implementation. $d_{\text{mesh}}(\alpha, \beta)$ measures the 
geometric consistency between detections $\alpha$ and $\beta$ 
by averaging the point-to-epipolar-line distances over all 
mesh vertices. Pairs with $d_{\text{mesh}}$ exceeding a threshold 
$\tau_{\text{mesh}}$ are rejected.
 
\noindent\textbf{Cost Matrix and Optimal Assignment.}
For all pairs $(\alpha, \beta)$ satisfying both filters, MAEM constructs a 
cost matrix $\mathbf{C}$ where each entry is the PA-MPJPE between the 3D 
keypoints of detection $\alpha$ in view $c_a$ and detection $\beta$ in view 
$c_b$:

\begin{equation}
    \mathbf{C}(\alpha, \beta) = \frac{1}{J} \sum_{j=1}^{J} 
    \left\| \mathbf{q}_{\alpha,j}^{c_a} - 
    \mathcal{T}\!\left(\mathbf{q}_{\beta,j}^{c_b}\right) \right\|_2
\end{equation}

\noindent where $\{\mathbf{q}_{\alpha,j}^{c_a}\}_{j=1}^{J}$ and 
$\{\mathbf{q}_{\beta,j}^{c_b}\}_{j=1}^{J}$ are the 3D keypoints of 
detections $\alpha$ and $\beta$ in their respective camera-relative 
coordinates, $\mathcal{T}(\cdot)$ denotes the Procrustes alignment (rotation, 
translation, and uniform scaling) of the second keypoint set onto the first, which minimises the residual per-joint distance between the two 
keypoint sets. The resulting cost matrix is passed to the Hungarian algorithm~\cite{kuhn1955hungarian}, producing a 
globally optimal one-to-one matching for each camera pair $(c_a, c_b)$.
 
\noindent\textbf{Multi-View Clustering via Union-Find.}
Pairwise matches from all $\binom{N}{2}$ camera pairs are merged into 
multi-view person clusters using a Union-Find data structure~\cite{galler1964improved}, a disjoint-set structure that supports two operations: \textit{Find}, 
which identifies the root representative of a set, and \textit{Union}, 
which merges two sets into one. Each detection is initially assigned to its 
own set. When two detections $(\alpha, \beta)$ are matched by the Hungarian 
algorithm, their sets are merged into a single cluster via the Union operation, 
grouping observations of the same person across views.

However, since Hungarian matching is performed independently for each camera 
pair, the resulting clusters may contain conflicting observations, that is, two detections from the same camera assigned to the same cluster, which is physically impossible, as a single individual can only appear once in a given camera view. To resolve such conflicts, an iterative procedure is applied: given a cluster containing two detections from the same camera $c$, the detection with the higher PA-MPJPE cost from the Hungarian assignment is removed as it is considered the less reliable match:

\begin{equation}
    \hat{d} = \arg\max_{d \in \mathcal{C}_c} \, c_{\text{PA}}(d),
\end{equation}

\noindent where $\mathcal{C}_c$ denotes the set of conflicting detections 
from camera $c$ within the cluster, and $c_{\text{PA}}(d)$ is the PA-MPJPE cost of detection $d$ recorded during Hungarian assignment. The detection $\hat{d}$ is removed from the cluster, and the procedure is repeated until no cluster contains more than one detection per camera.

\subsection{Per-Keypoint RANSAC Triangulation} \label{sec:triangulation}
Given the multi-view clusters from Stage~2, MAEM recovers the 3D position of each keypoint independently using a RANSAC~\cite{fischler1981random} robust triangulation scheme. The input for this stage is the set of projected 2D keypoints generated in Stage~1 and aggregated across all views within each cluster.

For each person cluster and each keypoint $j$, all $\binom{N_c}{2}$ view pairs 
$(c_a, c_b)$ within the cluster are used to generate candidate 3D hypotheses. 
Similarly to Step~A, the projected 2D keypoint locations $\mathbf{p}_{i,j}^{c_a}$ 
and $\mathbf{p}_{i,j}^{c_b}$ are triangulated via DLT to obtain a candidate 
3D point $\tilde{\mathbf{X}}_j^{(c_a,c_b)} \in \mathbb{R}^3$ in the world 
coordinate system. Each candidate is then reprojected into all $N_c$ views, and the number of inlier views is counted, where an inlier is defined as a view whose 
reprojection error falls below $\tau_{\text{reproj}}^{\text{kps}}$:

\begin{equation}
(c_a^*, c_b^*) = \arg\max_{(c_a, c_b)} 
\sum_{c=1}^{N_c}
\mathbbm{1}\!\left[
\left\|
\pi_c\!\left(\tilde{\mathbf{X}}_j^{(c_a,c_b)}\right)
- \mathbf{p}_{i,j}^{c}
\right\|_2
< \tau_{\text{reproj}}^{\text{kps}}
\right]
\end{equation}

\noindent where $\tilde{\mathbf{X}}_j^*$ is then defined as:

\begin{equation}
\tilde{\mathbf{X}}_j^* = \tilde{\mathbf{X}}_j^{(c_a^*, c_b^*)}
\end{equation}


\noindent where $\pi_c(\cdot)$ denotes the projection function of camera $c$. $(c_a^*, c_b^*)$ is the view pair that maximizes the number of consistent reprojections, and $\tilde{\mathbf{X}}_j^*$ denotes the corresponding best two-view hypothesis. $\tilde{\mathbf{X}}_j^*$ denotes the best two-view hypothesis. The final 
3D estimate $\mathbf{X}_j^*$ is obtained by re-triangulating keypoint $j$ 
using all inlier views $\mathcal{I}^* = \{c : \|\pi_c(\tilde{\mathbf{X}}_j^*) - \mathbf{p}_{i,j}^{c}\|_2 < \tau_{\text{reproj}}^{\text{kps}}\}$, where $|\mathcal{I}^*| \geq 2$, 
via confidence-weighted DLT across all views in $\mathcal{I}^*$.

\section{Experimental Settings}\label{Experiments}
\subsection{Dataset}\label{sec:dataset}
\begin{figure*}[t]
\centering

\subcaptionbox{Sportcenter EPFL\label{fig:epfl_1}}
  [0.32\textwidth]{\includegraphics[width=0.32\textwidth]{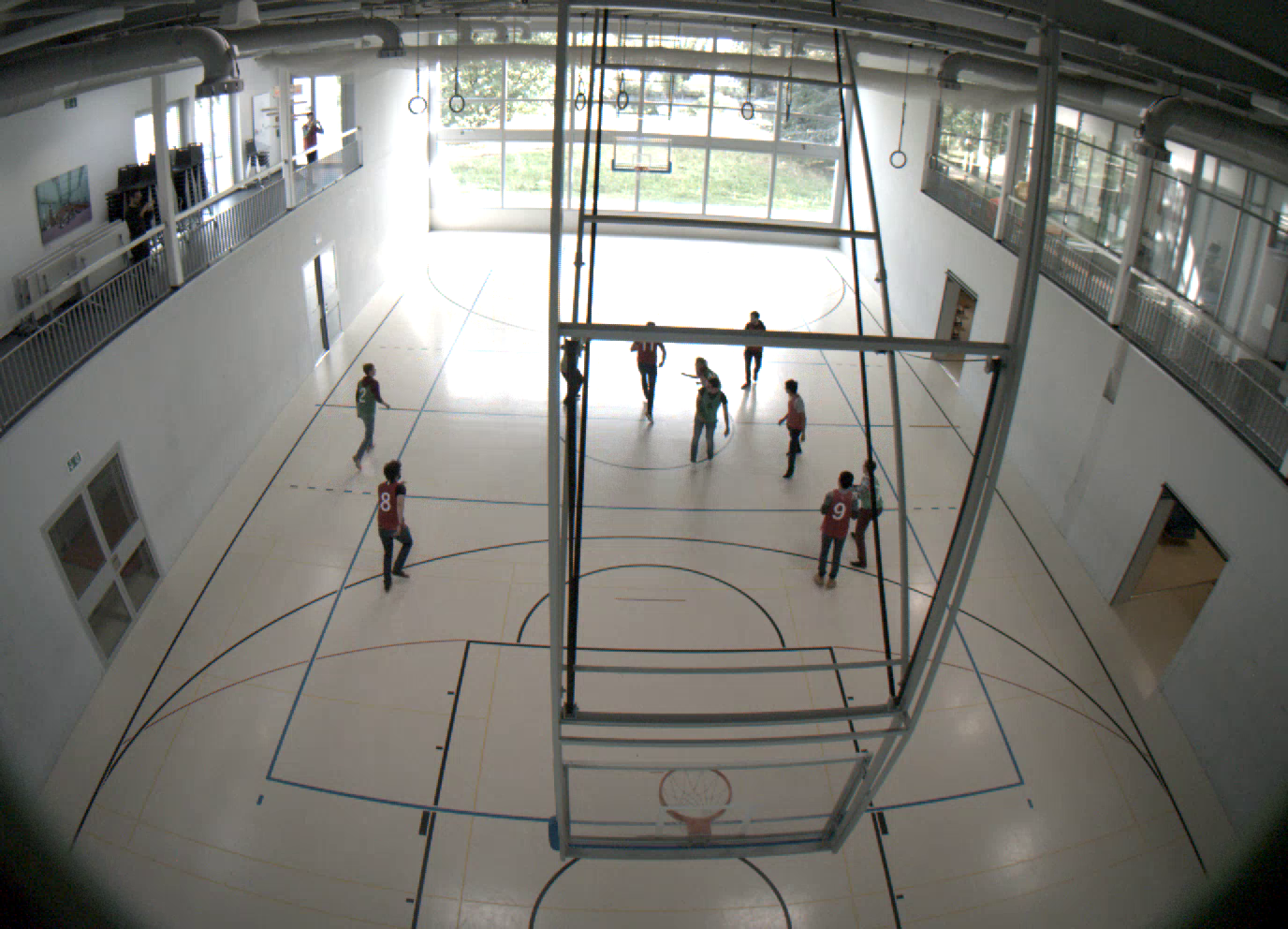}}
\hfill
\subcaptionbox{Sportcenter EPFL\label{fig:epfl_2}}
  [0.32\textwidth]{\includegraphics[width=0.32\textwidth]{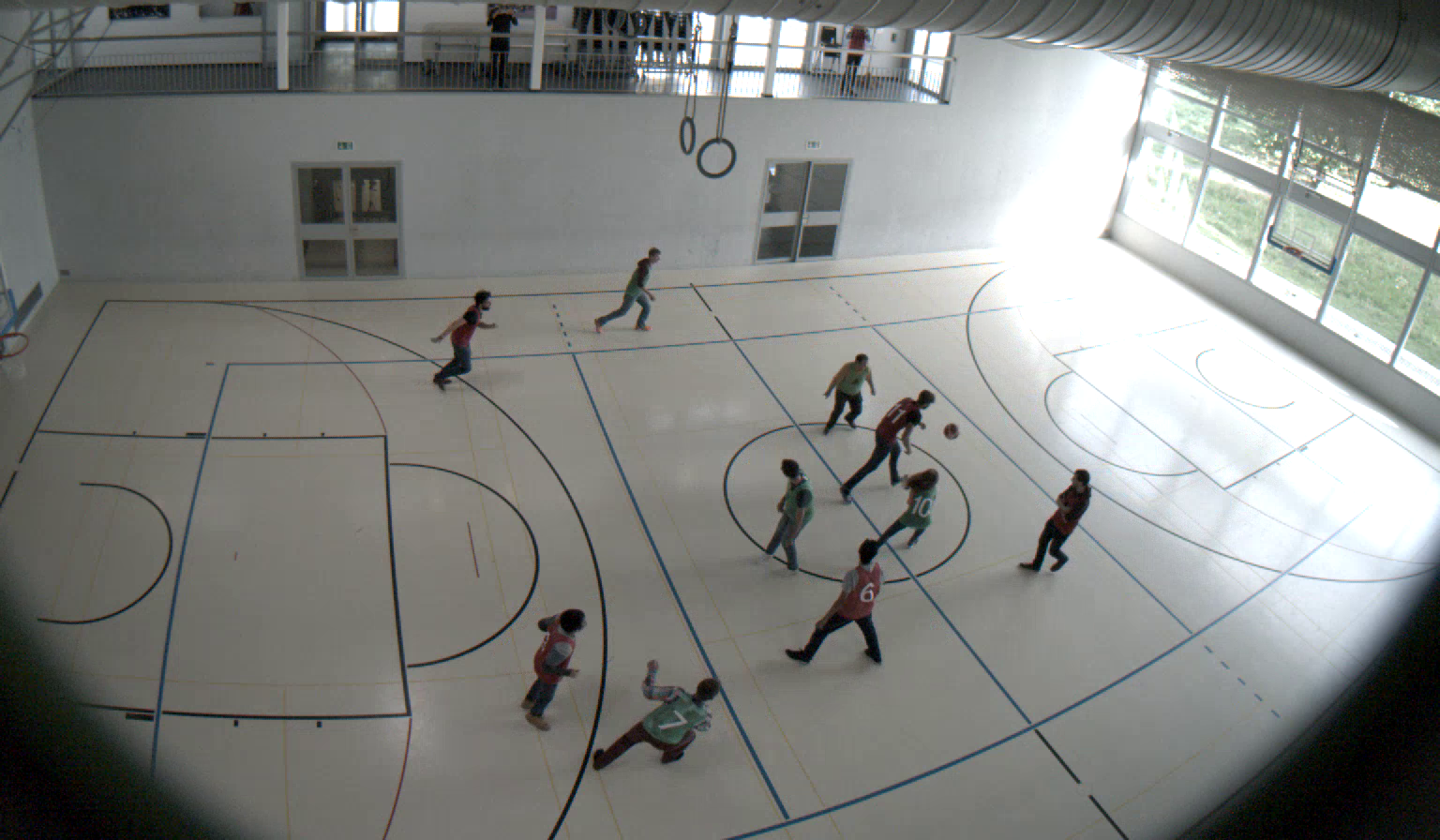}}
\hfill
\subcaptionbox{Sportcenter EPFL\label{fig:epfl_3}}
  [0.32\textwidth]{\includegraphics[width=0.32\textwidth]{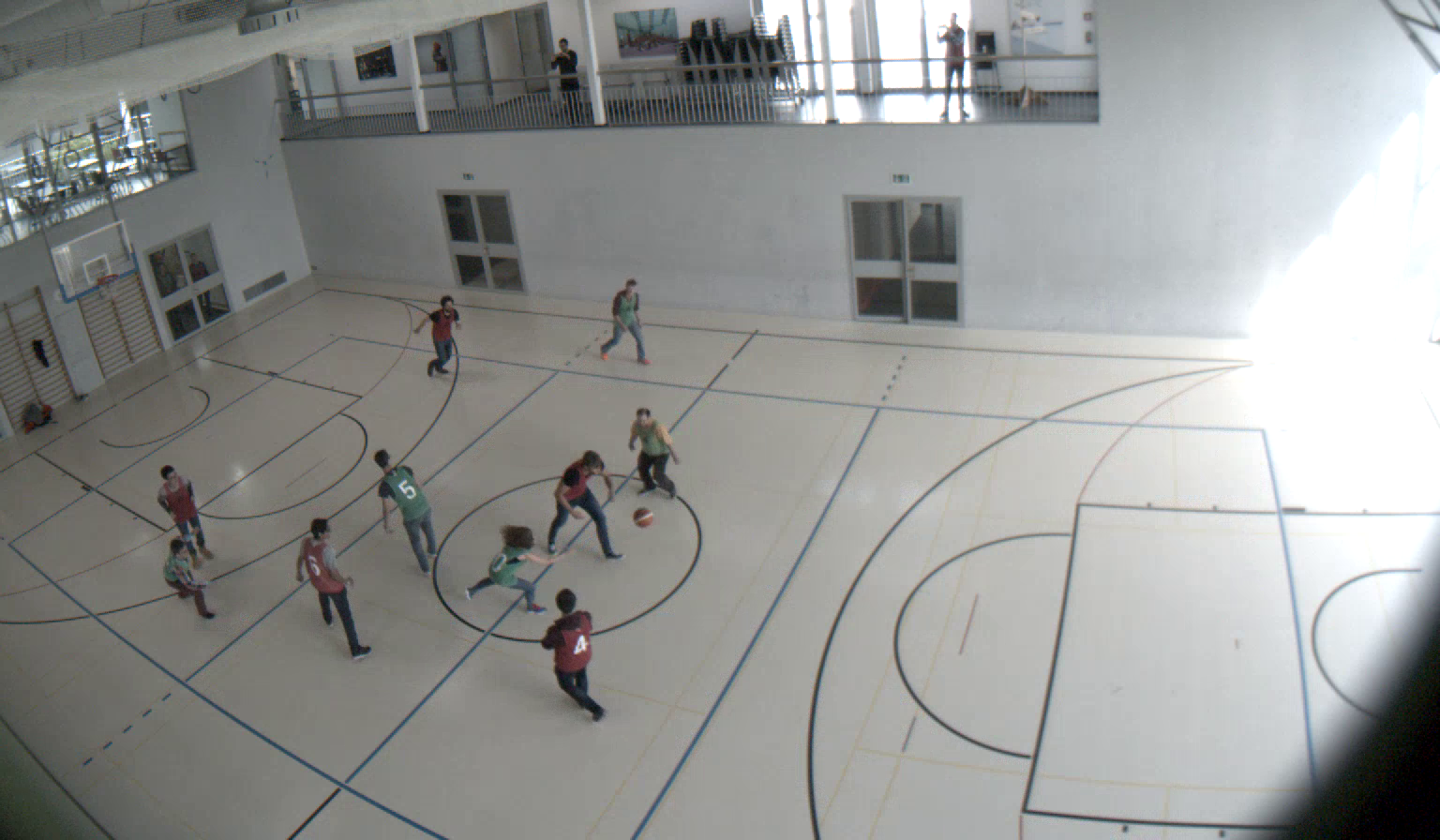}}

\vspace{6pt}

\subcaptionbox{Human-M3 Basketball1\label{fig:b1_1}}
  [0.24\textwidth]{\includegraphics[width=0.24\textwidth]{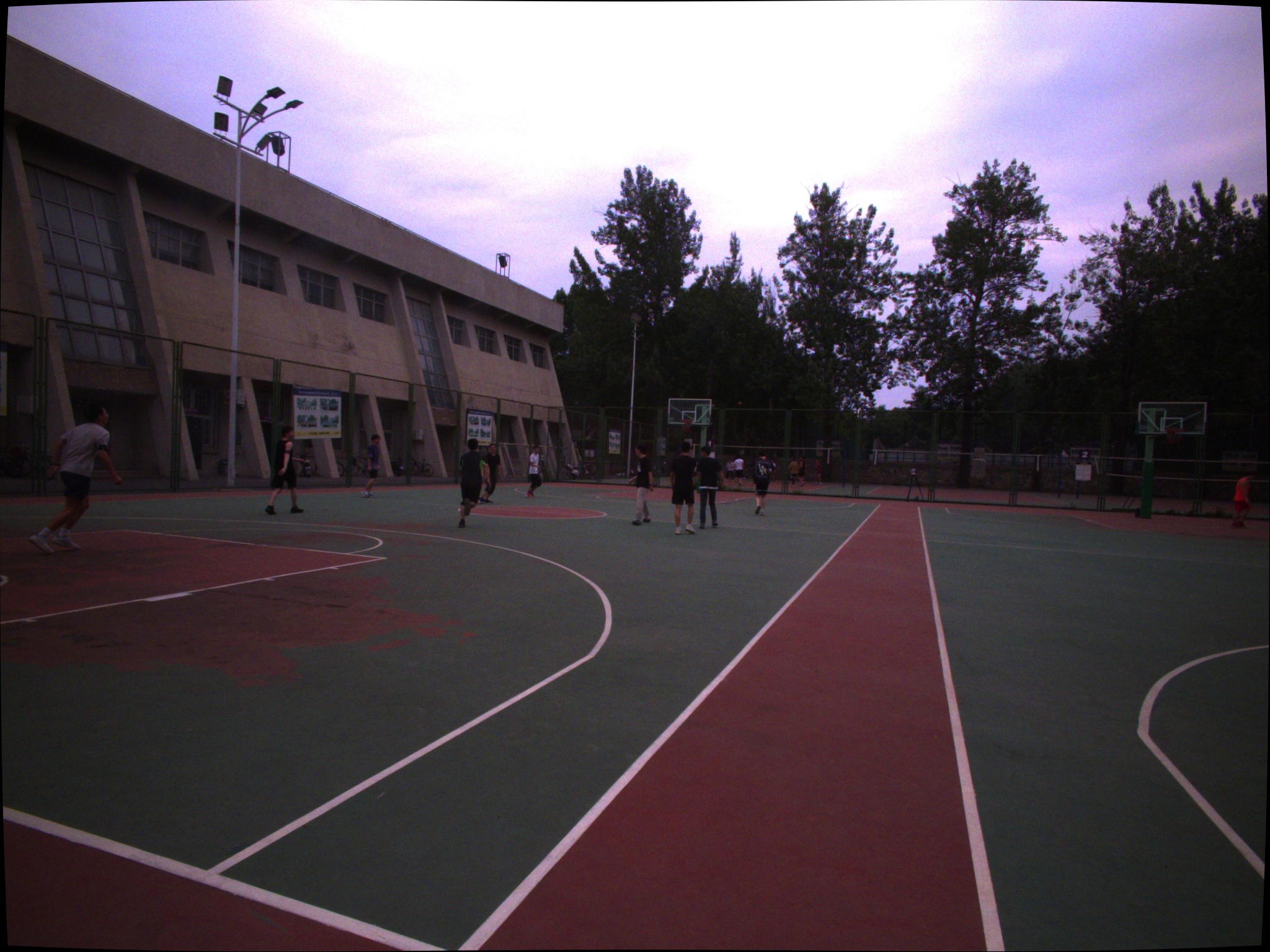}}
\hfill
\subcaptionbox{Human-M3 Basketball1\label{fig:b1_2}}
  [0.24\textwidth]{\includegraphics[width=0.24\textwidth]{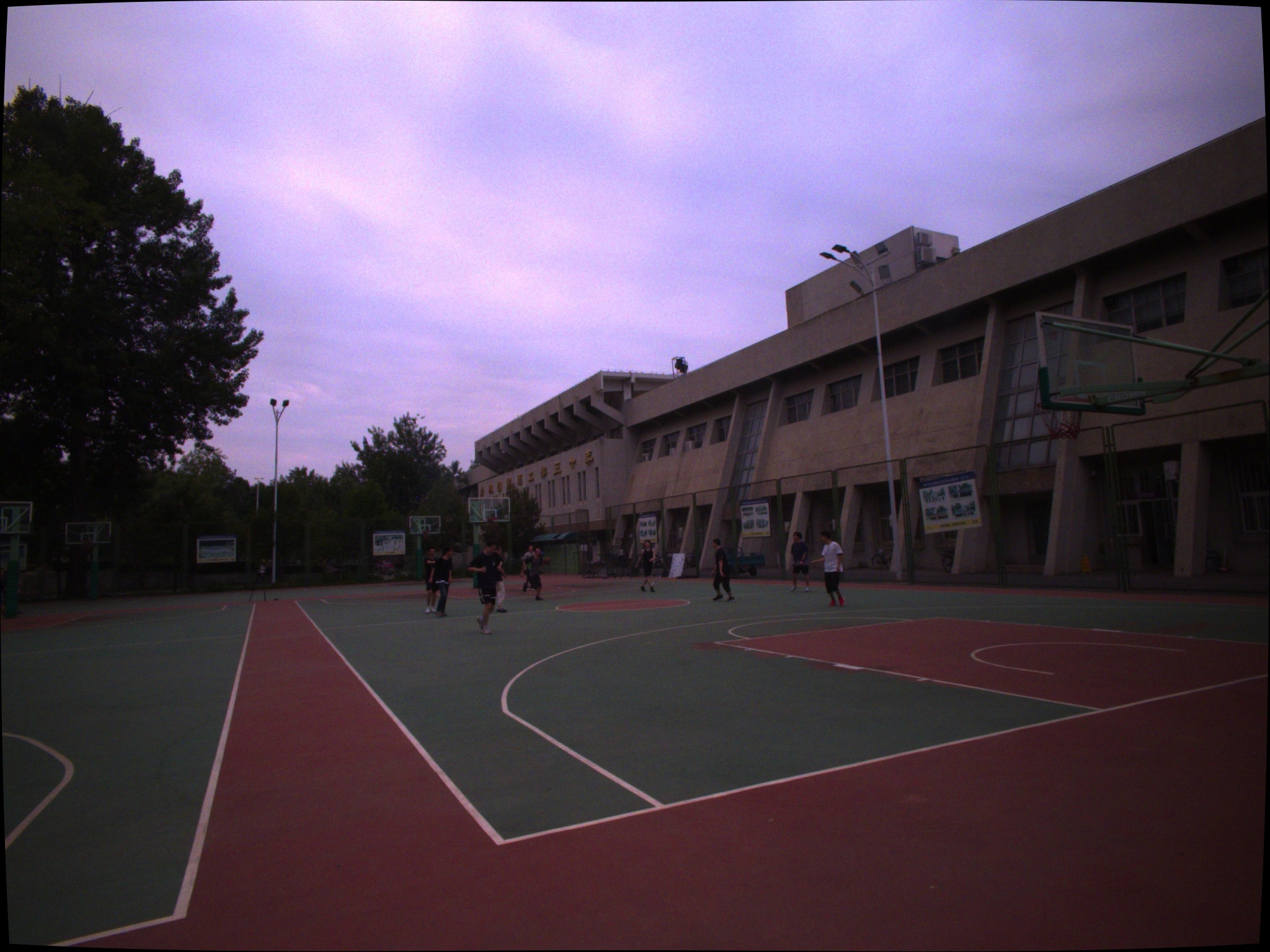}}
\hfill
\subcaptionbox{Human-M3 Basketball2\label{fig:b2_1}}
  [0.24\textwidth]{\includegraphics[width=0.24\textwidth]{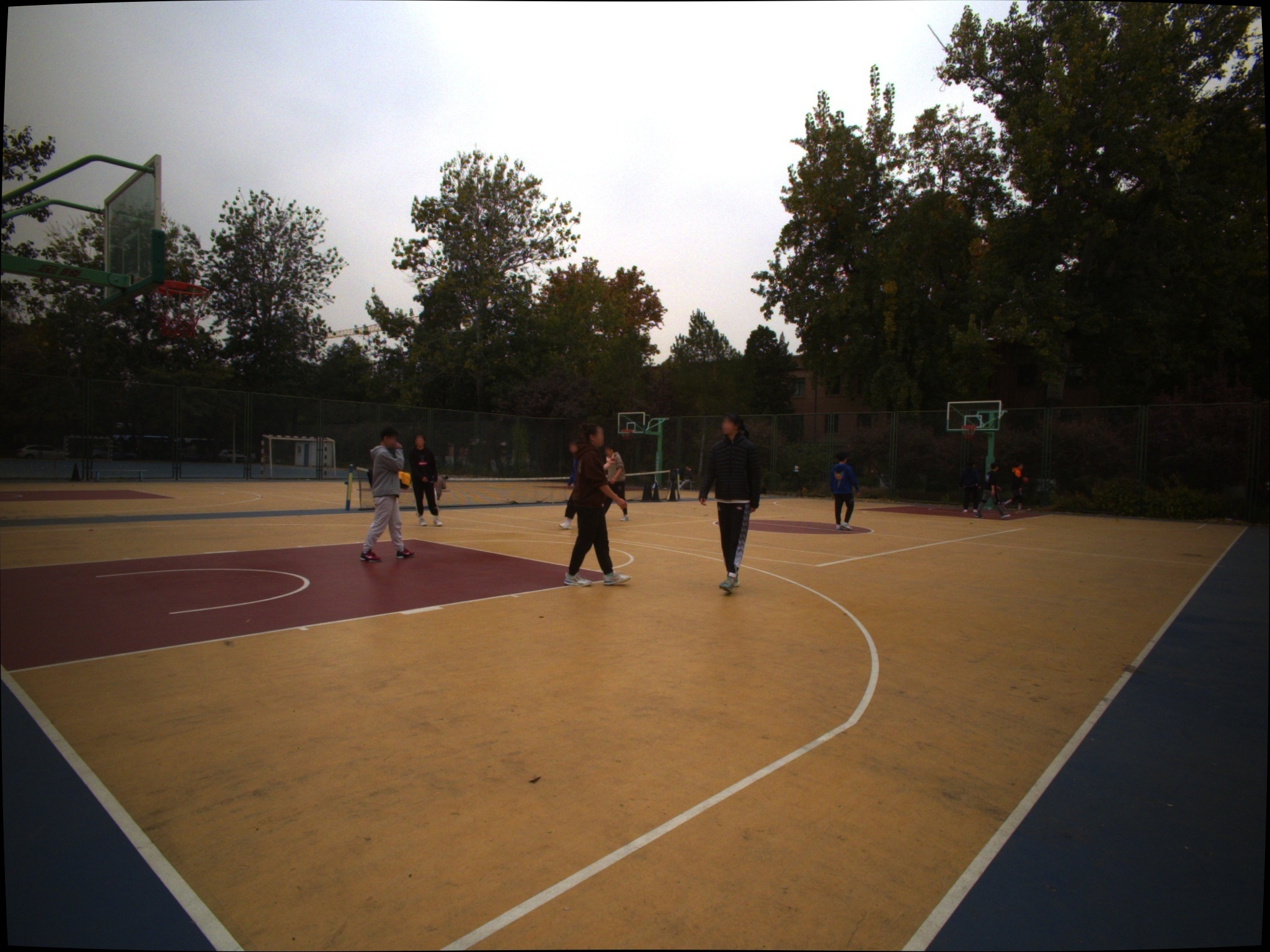}}
\hfill
\subcaptionbox{Human-M3 Basketball2\label{fig:b2_2}}
  [0.24\textwidth]{\includegraphics[width=0.24\textwidth]{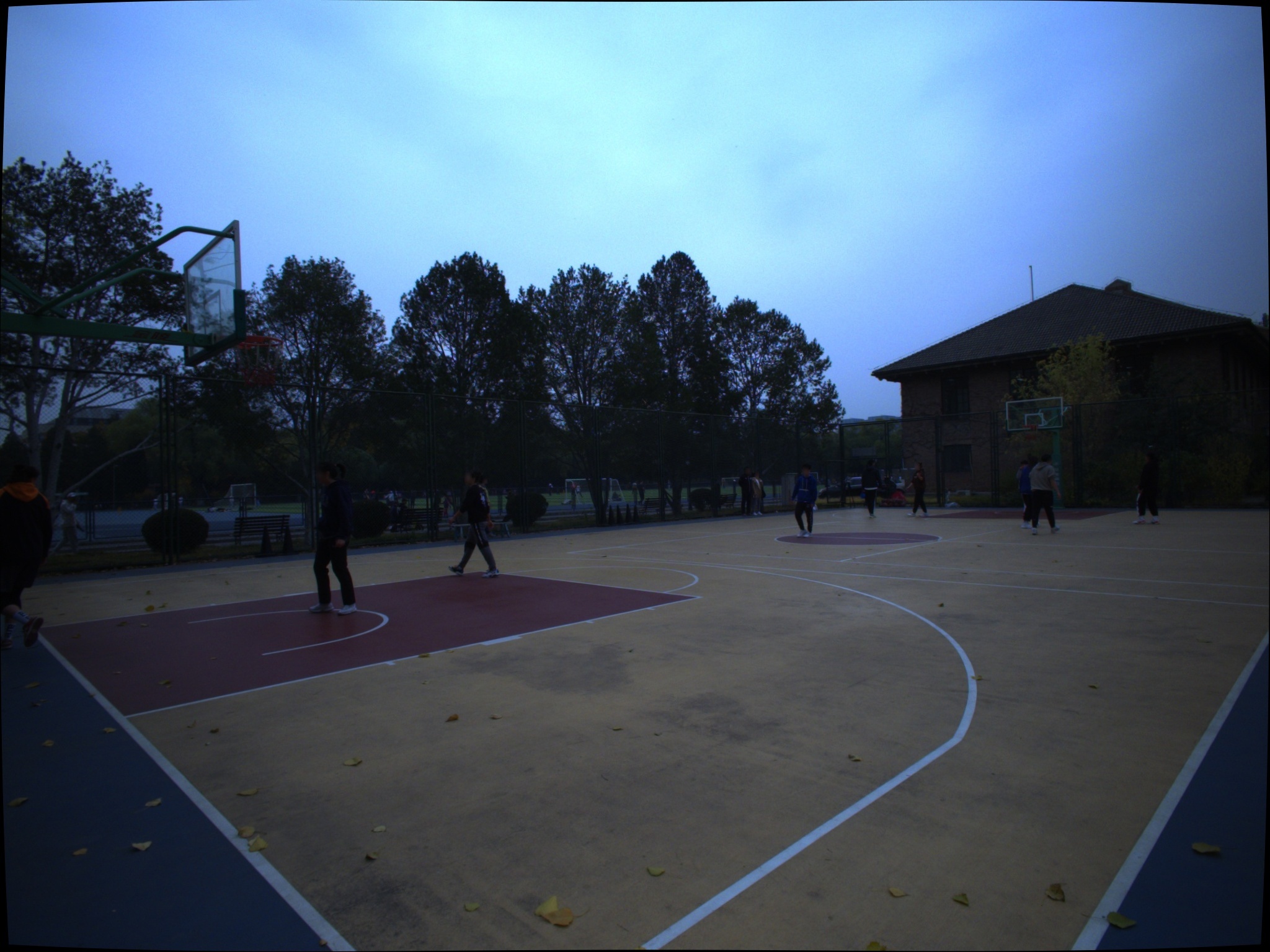}}

\caption{Example images from the evaluation datasets.
Top row: Sportcenter EPFL Multiview dataset, captured by elevated
fisheye cameras in an indoor gym.
Bottom row: Human-M3-Basketball sequences, captured by person-height
cameras mounted at court corners. Basketball1 uses 4 cameras and
Basketball2 uses 3 cameras.}
\label{fig:datasets}
\end{figure*}

To demonstrate the robustness of MAEM, we evaluated it on two public
multi-view amateur basketball datasets captured across diverse settings,
each featuring up to ten players on the court. Both datasets present
significant challenges for cross-view association, including frequent
occlusion, similar player appearances, and varying camera
configurations. Due to the wide-baseline camera placement and limited
field-of-view overlap, not every player is visible in all
camera views, further increasing the difficulty of
cross-view association. The key characteristics of both datasets are summarized in
Table~\ref{tab:dataset_summary}, and example images from
each dataset are shown in Figure~\ref{fig:datasets}.

\noindent\textbf{Sportcenter Dataset.}
This dataset~\cite{sportcenter_multiview} was collected in an indoor gym
with eight synchronized and calibrated fisheye cameras at 30~fps, each
operating at its own native resolution, ranging from 1328$\times$896 to
1536$\times$1200 pixels. The dataset provides 2D and 3D poses
annotations for 70 frames with 13 body keypoints per person. The 3D
ground truth was obtained by manually annotating 2D keypoints in each
view and triangulating them across views; as a result, some keypoints
are missing when they are occluded or not visible in a sufficient
number of views. In our experiments, we excluded the two roof-mounted top-view cameras and utilized the remaining six side-view cameras, as the extreme downward perspective of the top-view cameras led to significant person detection failure and severe body part occlusion.

\noindent\textbf{Human-M3 Dataset.} Human-M3~\cite{fan2023human} is a large-scale outdoor dataset with ground-truth 3D poses obtained through a multi-modal annotation pipeline combining multi-view RGB images with LiDAR point clouds. The dataset comprises multiple scenes, including basketball courts, a plaza, and an intersection. We evaluated exclusively on the two basketball sequences, referred to as Human-M3-Basketball. Specifically, Human-M3-Basketball consists of Basketball1, captured with 4 calibrated cameras across 400 test frames, and Basketball2, captured with 3 calibrated cameras across 500 test frames, both at 2048$\times$1536 resolution.

\begin{table}[htbp]
\centering
\caption{Summary of evaluation datasets.}
\label{tab:dataset_summary}
\small
\begin{tabular}{lcccc}
\hline
Dataset & Cameras used & Resolution & Frames & GT Poses \\
\hline
Sportcenter EPFL & 6 & 1328$\times$896 -- 1536$\times$1200 & 70 & 700 \\
Human-M3 Bball1 & 4 & 2048$\times$1536 & 400 & 3821 \\
Human-M3 Bball2 & 3 & 2048$\times$1536 & 500 & 4751 \\
\hline
\end{tabular}
\end{table}

\subsection{Baseline Methods}\label{sec:baseline methods}

\subsubsection{Training-Free Baselines}
\noindent\textbf{Baseline 1: MVPose.}
MVPose~\cite{dong2019fast} shares the same general pipeline as MAEM, namely per-view 2D pose estimation followed by cross-view matching and triangulation, but differs in its matching strategy. While MAEM relies purely on geometric constraints derived from calibrated cameras, MVPose establishes cross-view correspondences by combining geometric consistency with appearance similarity, formulated as a convex multi-way matching problem with a 
cycle-consistency constraint to produce globally consistent correspondences across all views. We implement this baseline using the XRMoCap toolbox~\cite{xrmocap}, with the original 2D pose estimator replaced by ViTPose~\cite{xu2022vitpose} to provide a more competitive front-end. Matched 2D keypoints were then triangulated to recover 3D poses. However, in team sports where players wear identical uniforms and are captured from long 
shooting distances, appearance features become unreliable, leading to frequent cross-view matching errors.

\noindent\textbf{Baseline 2: PA-MPJPE Matching.}
Unlike MAEM, which requires pre-calibrated cameras and uses geometric constraints for cross-view matching, Hokari~\emph{et~al.}~\cite{hokari2025human} propose a calibration-free cross-view matching approach that associates monocular mesh reconstructions across views based solely on pose similarity. However, since their coordinate unification relies on Procrustes alignment of root translations rather than camera calibration, the resulting 3D poses reside in an arbitrary coordinate system incompatible with ground-truth metrics defined in the calibrated world frame.

To enable full evaluation, we adopted only the cross-view association component of Hokari~\emph{et~al.}\ as a drop-in 
replacement within the MAEM pipeline. Specifically, cross-view matching was performed by constructing a complete graph over all per-view meshes with edges weighted by pairwise PA-MPJPE, and clustering them into per-person groups via spectral clustering. Once per-person clusters are formed, we replaced the original mesh fusion of Hokari~\emph{et~al.}\ with the same confidence-weighted triangulation as MAEM, where person detection 
confidence scores were used to weight the contribution of each view during 3D pose recovery. All remaining components, including the single-view front-end SAM 3D Body~\cite{yang2026sam3dbody}, triangulation, and post-processing, as well as all thresholds and parameters such as $\tau_{\text{det}}$ and $K_{\min}$, were kept 
identical to MAEM, as detailed in Table~\ref{tab:params}. This isolated the effect of the matching strategy alone.

\subsubsection{Existing Learning-Based Methods}

On the SportCenter dataset~\cite{sportcenter_multiview}, Roy~et~al.~\cite{roy2022triangulation} evaluated a semi-supervised approach that imposes multi-view geometric constraints via weighted differentiable triangulation. Using this formulation as self-supervision to train a 2D pose estimator and an auxiliary 3D lifting network, they reported an MPJPE of 66.9~mm. This result was obtained under a specific subject-based split, where the remaining subjects were used for training and only two subjects were reserved for evaluation. In contrast, MAEM is a training-free multi-view inference pipeline that requires no target-domain annotations. We evaluated on all annotated frames without partitioning them into a training split. Since the inference settings and evaluation subsets differ, directly comparing MPJPE values in the same table would be misleading. We therefore omitted their results from the quantitative comparison table.

On the Human-M3 dataset, we additionally compared MAEM against learning-based methods whose results were reported in the original benchmark~\cite{fan2023human}, including VoxelPose~\cite{tu2020voxelpose}, PlaneSweepPose~\cite{lin2021multi}, MVP~\cite{zhang2021direct}, and MMVP~\cite {fan2023human}. Note that MMVP took both RGB images and LiDAR point clouds as input, whereas MAEM used only RGB images; comparison with MMVP, therefore, reflected a modality disadvantage for our method. These learning-based methods are trained on the Human-M3 training set and require labeled data from the target scene. By contrast, MAEM operates in a fully training-free manner and is applied directly to the test set without any fine-tuning. Since MAEM targets team-sport scenarios, we evaluated only on the Human-M3-Basketball. For MMVP, per-scene results on Human-M3-Basketball are available in the same benchmark and are directly comparable. For VoxelPose, PlaneSweepPose, and MVP, only overall results across Human-M3 are reported; we included these as a reference, noting that their evaluation encompasses non-basketball scenes. This highlights a key advantage of MAEM: as a training-free framework, it can be directly applied to scenarios where large-scale domain-specific annotations are unavailable.

\subsection{Evaluation Metrics}\label{sec:metric}

We adopted four standard metrics to evaluate the accuracy and completeness of 3D pose estimation.

\noindent\textbf{MPJPE} (Mean Per-Joint Position Error) measures the average Euclidean distance (in mm) between the predicted and ground-truth 3D joint positions without root alignment in the world coordinate system, measuring per-joint localization accuracy.

\noindent\textbf{PA-MPJPE} (Procrustes-Aligned Mean Per-Joint Position Error) first aligns the predicted pose to the ground-truth via Procrustes analysis, then computes the mean per-joint distance (in mm). By removing global pose differences, this metric isolates the accuracy of the estimated body configuration.

\noindent\textbf{Recall} is defined as the ratio of true positives to the total number of ground-truth instances. A predicted pose is considered a true positive if its MPJPE to the nearest unmatched ground-truth pose falls below a predefined threshold of 500~mm. This metric evaluates the ability of the pipeline to detect and reconstruct all persons present in the scene.

\noindent\textbf{AP$_{\delta}$} (Average Precision at threshold
$\delta$\,mm) measures the proportion of correctly detected persons
under varying strictness levels. A predicted pose is considered a true
positive if its MPJPE to the matched ground-truth falls below
$\delta$\,mm.

\subsection{Implementation Details}\label{sec:details}

\noindent\textbf{Single-View Human Mesh Recovery.}
We employed SAM 3D Body \cite{yang2026sam3dbody} as the monocular human mesh recovery model (Section~\ref{sec:single_view}), which requires no training or fine-tuning on the target dataset.

\noindent\textbf{Cross-View Matching and Clustering.}
Several thresholds govern the matching pipeline: detection confidence filtering, geometric filtering stringency, and the minimum number of views required to form a valid person cluster, each affecting the precision-recall trade-off of the final matching result. These thresholds were tuned per dataset to accommodate differences in camera count, mounting height, and imaging conditions, as summarized in Table~\ref{tab:params}.

\begin{table}[t]
\centering
\caption{Dataset-specific thresholds used in MAEM. Note: px denotes pixels.}
\label{tab:params}
\begin{tabular}{lcc}
\toprule
Parameter & Sportcenter EPFL & Human-M3 \\
\midrule
Detection confidence  $\tau_{\text{det}}$ & 0.9 & 0.7 \\
Reprojection threshold  $\tau_{\text{reproj}}^{\text{bbox}}$ & 10.0~px & 30.0~px \\
Mesh Epipolar threshold  $\tau_{\text{mesh}}$ & 8.0~px & 8.0~px \\
Min views per cluster  $K_{\min}$ & 4 & 2 \\
RANSAC inlier threshold $\tau_{\text{reproj}}^{\text{kps}}$ & 20.0~px & 20.0~px \\
\bottomrule
\end{tabular}
\end{table}

The detection confidence threshold $\tau_{\text{det}}$ was set higher for Sportcenter EPFL to filter out non-player bystanders captured by the elevated fisheye cameras, and lower for Human-M3 to retain detections of heavily occluded players. The reprojection threshold $\tau_{\text{reproj}}^{\text{bbox}}$ was set to 10.0~px for Sportcenter EPFL, where six-camera redundancy supports aggressive geometric filtering, and relaxed to 30.0~px for Human-M3-Basketball, where person-height cameras and severe inter-player occlusion produce noisier bounding box centre estimates. The minimum cluster size $K_{\min}$ was set to 2 for Human-M3, the lowest permissible value, accepting a higher risk of geometrically under-constrained clusters in exchange for improved recall.

The epipolar threshold $\tau_{\text{mesh}}$ and the RANSAC inlier threshold $\tau_{\text{reproj}}^{\text{kps}}$ were kept consistent across both datasets at 8.0~px and 20.0~px, respectively. Unlike $\tau_{\text{reproj}}^{\text{bbox}}$, which depends on camera geometry and occlusion conditions, $\tau_{\text{mesh}}$ measures the mean point-to-epipolar-line distance over the $N_v$ projected MHR surface vertices. Since this quantity is grounded in the fixed mesh topology rather than scene-level geometric factors, its distribution remains stable across different camera configurations.


\noindent\textbf{Evaluation Protocol.}
Predicted 3D poses are matched to ground-truth poses via Hungarian assignment. A prediction is considered a false positive if its MPJPE to the matched ground-truth exceeds 500~mm.

\noindent\textbf{Hardware.}
All experiments were conducted on an NVIDIA RTX A6000 with 48~GB memory.

\section{Experiments and Results}\label{Results}
\subsection{Comparison with Baselines}\label{sec:baselines}


\begin{table}[htbp]
\centering
\footnotesize
\setlength{\tabcolsep}{4pt}
\caption{Performance comparison of MAEM against baselines on the Sportcenter EPFL Multiview datasets. For MPJPE and PA-MPJPE, results are reported in the format of Mean / Median.}
\label{tab:comparison_results}
\begin{tabular}{lccccccc}
\hline
\multirow{2}{*}{\textbf{Method}} & \textbf{Recall} & \textbf{MPJPE} & \textbf{PA-MPJPE} & \multicolumn{4}{c}{\textbf{AP $\uparrow$ (\%)}} \\
 & $\uparrow$ (\%) & $\downarrow$ (mm) & $\downarrow$ (mm) & \textbf{@75} & \textbf{@100} & \textbf{@125} & \textbf{@150} \\
\hline
MVPose \cite{dong2019fast} & --- & 2802.5 / 1789.7 & 1787.2 / 462.9 & --- & --- & --- & --- \\
PA-MPJPE Matching & 86.3 & 65.7 / 54.1 & 45.3 / 38.8 & 83.7 & 84.7 & 85.1 & 85.4 \\
\textbf{MAEM (Ours)} & \textbf{99.6} & \textbf{59.8 / 53.1} & \textbf{40.7 / 38.2} & \textbf{96.7} & \textbf{98.0} & \textbf{98.2} & \textbf{98.4} \\
\hline
\multicolumn{8}{l}{\footnotesize $^1$ Recall and AP for MVPose are omitted due to extreme localization drift (MPJPE > 2.5m).} \\
\end{tabular}
\end{table}

\begin{table}[htbp]
\centering
\caption{ Performance comparison on the Human-M3-Basketball dataset. MMVP utilizes RGB and PCD (point cloud data, from a LiDAR sensor), \textbf{requiring additional depth sensor input}, whereas all other methods use RGB only. For MPJPE and PA-MPJPE, results are reported as Mean / Median in mm.} 

\label{tab:humanm3_results}
\footnotesize
\setlength{\tabcolsep}{2pt}
\begin{tabular}{llcccc@{\hspace{4pt}}c@{\hspace{4pt}}c@{\hspace{4pt}}c}
\hline
\multirow{2}{*}{\textbf{Method}} & \multirow{2}{*}{\textbf{Input}} & \textbf{Recall} & \textbf{MPJPE} & \textbf{PA-MPJPE} & \multicolumn{4}{c}{\textbf{AP $\uparrow$ (\%)}} \\
 & & $\uparrow$ (\%) & $\downarrow$ (mm) & $\downarrow$ (mm) & \textbf{@75} & \textbf{@100} & \textbf{@125} & \textbf{@150} \\
\hline
MVPose \cite{dong2019fast} & RGB & --- & 4021.0 / 3393.0 & 1010.0 / 388.0 & --- & --- & --- & --- \\
PA-MPJPE Match. & RGB & 60.9 & 91.5 / 62.2 & 63.5 / 44.5 & 51.6 & 54.7 & 55.8 & 56.8 \\
MMVP \cite{fan2023human} & RGB+PCD & \textbf{98.4} & 75.9 / --- & --- / --- & 43.6 & 71.8 & \textbf{84.1} & \textbf{90.5} \\
\textbf{MAEM (Ours)} & RGB & 83.7 & \textbf{74.0 / 58.4} & \textbf{51.8 / 42.4} & \textbf{76.6} & \textbf{79.8} & 81.0 & 81.6 \\
\hline
\multicolumn{9}{l}{\footnotesize $^1$ Recall and AP for MVPose are omitted due to extreme localization drift (MPJPE > 3m).} \\
\multicolumn{9}{l}{\footnotesize $^2$ ``---'' for other methods indicates metrics unavailable in their original publications.} \\
\end{tabular}
\end{table}

Table~\ref{tab:comparison_results} reports the performance comparison on the SportCenter dataset. MAEM achieves the best performance across all metrics, attaining a Recall of 99.6\%, an MPJPE of 59.8/53.1~mm, and a PA-MPJPE of
40.7/38.2~mm. Compared to PA-MPJPE Matching, Recall improves from 86.3\% to 99.6\%, indicating that geometry-based cross-view constraints allow MAEM to associate a greater number of players across views, including cases where pose-similarity-based matching
struggles due to ambiguous configurations among nearby players. The AP scores further
confirm the precision advantage of MAEM: AP@75 improves from 83.7\% to 96.7\% and
AP@100 from 84.7\% to 98.0\%, suggesting that the recovered 3D poses are not only
more complete in terms of person coverage, but also more accurately localized in 3D
space. The gap widens further under stricter thresholds, with AP@125 improving from
85.1\% to 98.2\% and AP@150 from 85.4\% to 98.4\%, collectively demonstrating that
MAEM produces consistently precise pose estimates across a wide range of localization
tolerances. MVPose produces an extreme MPJPE exceeding 2800~mm with no valid AP scores reported, indicating a fundamental failure of appearance-based cross-view matching under these conditions.

Supplementary Table S1 presents results on the full Human-M3 dataset, reproduced in~\cite{fan2023human}, where methods were trained and evaluated following the train/test split defined by the dataset. Note that these results cover all scenes in Human-M3, including non-basketball sequences, and are therefore not directly comparable to the basketball-only evaluation in Table~\ref{tab:humanm3_results}. MMVP achieves the best performance among all compared learning-based methods, with the lowest MPJPE of 79.0 mm and the highest AP$_{75}$ of 41.9\% and AP$_{100}$ of 69.9\%, providing contextual reference for the broader Human-M3 benchmark.

Table~\ref{tab:humanm3_results} reports results on the Human-M3-Basketball 
dataset. The table presents results on the test split of Human-M3-Basketball, 
where MAEM, MVPose, and PA-MPJPE Matching were evaluated. Among RGB-only methods on Human-M3-Basketball, MAEM achieves 
an MPJPE of 74.0/58.4~mm and PA-MPJPE of 51.8/42.4~mm, 
outperforming PA-MPJPE Matching on all metrics. AP scores 
improve consistently across all thresholds, from 51.6\% to 
76.6\% at @75, 54.7\% to 79.8\% at @100, 55.8\% to 81.0\% 
at @125, and 56.8\% to 81.6\% at @150. MVPose again fails 
to produce valid results, with MPJPE exceeding 4000~mm and 
no AP scores reported. Compared to MMVP, which uses additional 
LiDAR input, MAEM achieves lower Recall (83.7\% vs.\ 98.4\%) 
but superior AP at stricter thresholds, with AP@75 of 76.6\% 
vs.\ 43.6\% and AP@100 of 79.8\% vs.\ 71.8\%, using RGB only.

\begin{figure}[htbp]
    \centering
    \setlength{\tabcolsep}{2pt}
    \begin{tabular}{cccccccc}
    \includegraphics[width=0.11\linewidth]{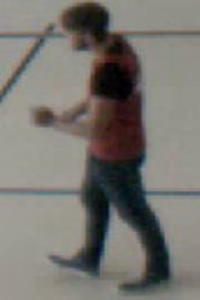} &
    \includegraphics[width=0.11\linewidth]{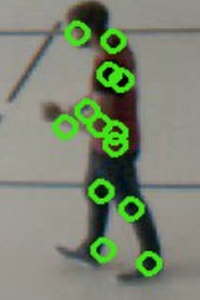} &
    \includegraphics[width=0.11\linewidth]{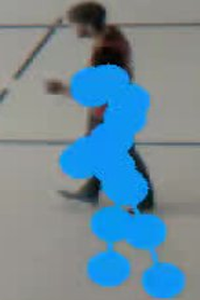} &
    \includegraphics[width=0.11\linewidth]{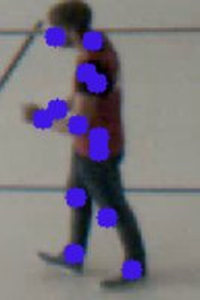} &
    \includegraphics[width=0.11\linewidth]{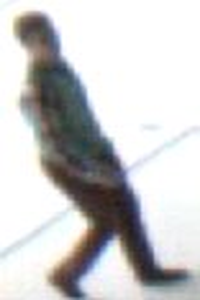} &
    \includegraphics[width=0.11\linewidth]{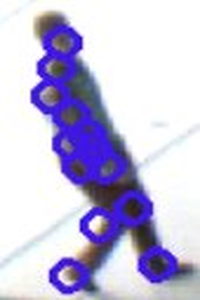} &
    \includegraphics[width=0.11\linewidth]{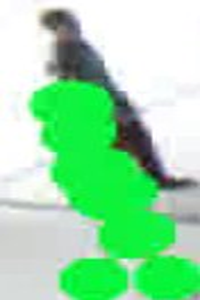} &
    \includegraphics[width=0.11\linewidth]{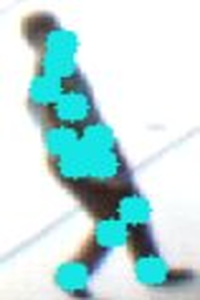} \\[2pt]
    \includegraphics[width=0.11\linewidth]{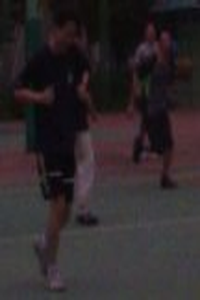} &
    \includegraphics[width=0.11\linewidth]{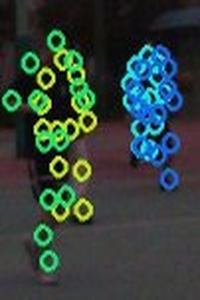} &
    \includegraphics[width=0.11\linewidth]{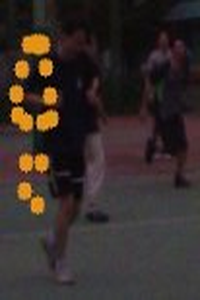} &
    \includegraphics[width=0.11\linewidth]{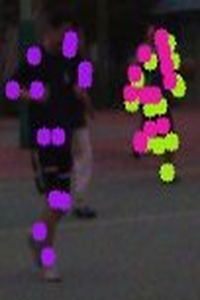} &
    \includegraphics[width=0.11\linewidth]{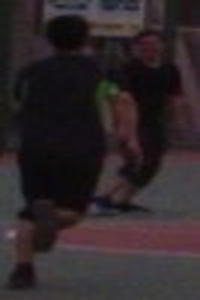} &
    \includegraphics[width=0.11\linewidth]{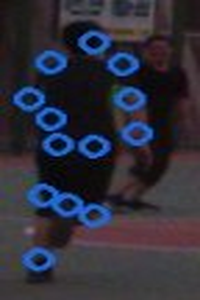} &
    \includegraphics[width=0.11\linewidth]{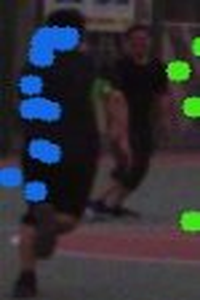} &
    \includegraphics[width=0.11\linewidth]{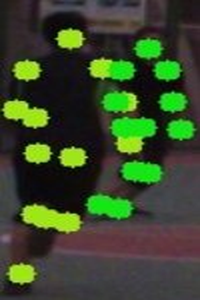} \\[2pt]
    \includegraphics[width=0.11\linewidth]{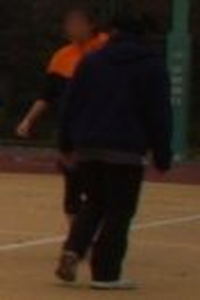} &
    \includegraphics[width=0.11\linewidth]{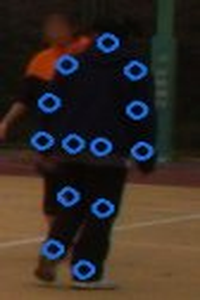} &
    \includegraphics[width=0.11\linewidth]{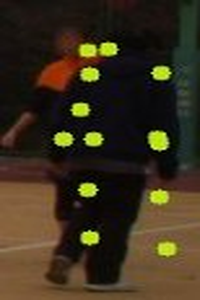} &
    \includegraphics[width=0.11\linewidth]{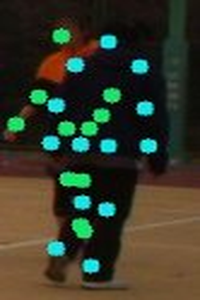} &
    \includegraphics[width=0.11\linewidth]{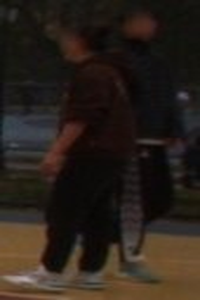} &
    \includegraphics[width=0.11\linewidth]{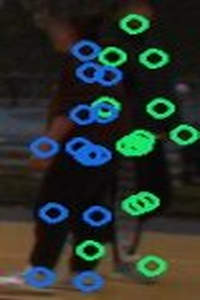} &
    \includegraphics[width=0.11\linewidth]{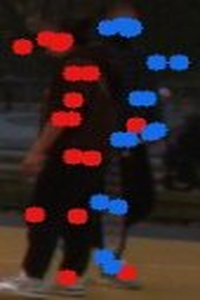} &
    \includegraphics[width=0.11\linewidth]{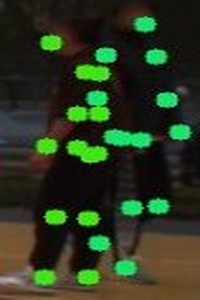} \\[2pt]
    Input & GT & MVPose & MAEM &
    Input & GT & MVPose & MAEM \\
\end{tabular}
    \caption{Qualitative results on the two datasets. Each row shows, from left to right, 
    the input image, ground truth, MVPose prediction, and MAEM prediction. 
    Rows correspond to Sportcenter EPFL (top), Human-M3 Basketball1 (middle), 
    and Human-M3 Basketball2 (bottom).}
    \label{fig:results_sample} 
\end{figure}

Figure~\ref{fig:results_sample} illustrates the qualitative 3D pose results of MAEM on the two datasets, where the estimated 3D poses are reprojected onto the image planes. Each row displays, from left to right, the input image, ground truth, MVPose prediction, and MAEM prediction. In the ground truth column, each color represents a distinct person identity. Different colors within the same image indicate different individuals. Compared to MVPose, MAEM produces more accurate and complete pose estimates, particularly in crowded scenes with severe occlusion, as shown in the Human-M3-Basketball sequences (middle and bottom rows). In the middle row, one person present in the ground truth is absent from the MAEM prediction, typically due to severe occlusion limiting visibility to fewer than $K_{min}$ views. Additional qualitative results are provided in Supplementary Figure S2.

\subsection{Ablation Studies}\label{sec:ablation}

We conducted ablation experiments on the Human-M3 \cite{fan2023human} dataset to analyze the contribution of each component in the matching pipeline. Results were summarized in
Table~\ref{tab:ablation_matching}.

The reprojection filter alone achieves a Recall of only 61.4\% with an
MPJPE of 95.8~mm, indicating that it can eliminate spatially implausible
pairs but lacks the discriminability to resolve ambiguities among nearby players.

The dense mesh epipolar filter alone improves Recall to 75.1\% and MPJPE to 79.2~mm, but remains susceptible to degenerate cases: when epipolar lines are approximately parallel to the player arrangement on the court, players at different spatial locations may produce similar epipolar distances, leading to incorrect associations. The reprojection filter addresses this by first eliminating geometrically implausible pairs in 3D, ensuring that only spatially consistent candidates enter the dense epipolar verification.

Replacing the dense mesh vertices with 15 sparse keypoints in the epipolar filter, while retaining the reprojection filter, leads to degraded performance across all metrics, with a particularly pronounced drop in Recall from 83.7\% to 47.5\%. This sharp decline stems from two compounding factors. First, the limited camera count (3--4 views) means that a single erroneous rejection at the epipolar filtering stage may eliminate the only geometrically valid correspondence for a given player, directly causing that player to fall below $K_{\min}$ and be discarded. Second, severe inter-player occlusion at person-height viewpoints frequently truncates the visible body region, leaving too few sparse keypoints reliably projected to produce a discriminative epipolar consistency signal. The dense mesh mitigates both failure modes by distributing geometric evidence across $N_v$ surface vertices, ensuring a stable estimate even under heavy occlusion. These results confirm that feature density is a structural necessity 
for robust recall under limited camera coverage and severe occlusion, rather than merely a precision refinement.

The full MAEM pipeline combines both the reprojection filter and the dense mesh epipolar filter, achieving the best results across all metrics: 83.7\% Recall, 74.0~mm MPJPE, and 76.6\% AP$_{75}$. This confirms that the two filtering steps are complementary: the reprojection filter efficiently reduces the candidate space, while the dense mesh epipolar filter provides fine-grained geometric verification that sparse keypoints cannot match.

\begin{table}[htbp]
\centering
\caption{Ablation study on the matching mechanisms. "Reproj." denotes the bounding box reprojection filter. "Epipolar" indicates the type of features used for epipolar consistency check (None, Sparse Keypoints, or Dense Mesh). The full MAEM pipeline utilizes both the reprojection filter and the dense mesh epipolar filter. MPJPE and PA-MPJPE report mean values in mm.}
\label{tab:ablation_matching}
\small
\setlength{\tabcolsep}{3pt}
\begin{tabular}{cllccccc}
\toprule
\multirow{2}{*}{\textbf{Variant}} & \textbf{Reproj.} & \textbf{Epipolar} & \textbf{Recall} & \textbf{MPJPE} & \textbf{PA-MPJPE} & \multicolumn{2}{c}{\textbf{AP $\uparrow$ (\%)}} \\
\cmidrule(lr){7-8}
 & \textbf{Filter} & \textbf{Filter} & $\uparrow$ (\%) & $\downarrow$ (mm) & $\downarrow$ (mm) & \textbf{@75} & \textbf{@100} \\
\midrule
(a) & \checkmark & --- & 61.4 & 95.8 & 67.5 & 50.6 & 53.9 \\
(b) & --- & Dense Mesh & 75.1 & 79.2 & 54.9 & 67.2 & 70.3 \\
(c) & \checkmark & Sparse Kpts. & 47.5 & 74.2 & 53.3 & 43.5 & 45.1 \\
\midrule
\textbf{(d) MAEM} & \checkmark & \textbf{Dense Mesh} & \textbf{83.7} & \textbf{74.0} & \textbf{51.8} & \textbf{76.6} & \textbf{79.8} \\
\bottomrule
\end{tabular}
\end{table}

\subsection{Computational Efficiency}\label{sec:efficiency}

We evaluated the runtime of the two-stage cross-view matching on both datasets using a single CPU thread on an AMD EPYC 7302 16-Core Processor (3.0 GHz). Both datasets have a similar detection load of approximately 12 persons per view per frame. Human-M3-Basketball uses 4 cameras, resulting in 6 view pairs, while Sportcenter EPFL uses 6 cameras, resulting in 15 view pairs.

Table~\ref{tab:runtime} reports the average per-frame matching time. Despite having 2.5$\times$ more view pairs, Sportcenter EPFL achieves lower per-frame matching time than Human-M3-Basketball. This is because the wide camera baselines in Sportcenter EPFL make the reprojection filter highly selective (71.5\% rejection rate), substantially reducing the number of candidate pairs that reach the more expensive dense mesh epipolar filter. Since the epipolar
filter dominates runtime due to the 18,439-vertex distance computation, the overall cost scales with the number of surviving pairs rather than total pairs. This cascade design ensures that the method remains practical for multi-camera scenes with dense player detections.

\begin{table}[htbp]
\centering
\caption{Per-frame matching time on both datasets (single CPU thread).}
\label{tab:runtime}

\renewcommand{\arraystretch}{1.5} 
\setlength{\tabcolsep}{18pt}      
\small                            

\begin{tabular}{lccc}
\toprule  
Dataset & Cameras & View pairs & Time (ms) \\
\midrule  
Human-M3-Basketball & 4 & 6  & 1819.9 \\
Sportcenter EPFL    & 6 & 15 & 1427.2 \\
\bottomrule 
\end{tabular}
\end{table}

\section{Discussion}\label{sec:discussion}

The experimental results demonstrate that MAEM achieves robust multi-view multi-person 3D pose estimation across diverse team sports scenarios.

A direct comparison between MAEM and MMVP warrants careful interpretation, as the two methods operate under fundamentally different sensing assumptions. MMVP augments RGB images with LiDAR point clouds, which provide metric-scale depth measurements and dense geometric evidence largely invariant to occlusion and lighting. MAEM, by contrast, relies exclusively on RGB input, recovering depth indirectly through the shape and pose priors encoded in the monocular mesh recovery model. The Recall gap at loose thresholds (MMVP 98.4\% vs.\ MAEM 83.7\%) is therefore partly structural rather than purely algorithmic: depth sensors provide additional geometric cues that aid detection under challenging viewing conditions, whereas monocular mesh recovery depends solely on visible image evidence and is more sensitive to occlusion. At stricter thresholds, however, MAEM outperforms MMVP (AP@75: 76.6\% vs.\ 43.6\%; AP@100: 79.8\% vs.\ 71.8\%), suggesting that dense mesh projections provide finer-grained spatial localisation than the LiDAR-fused representation. These results indicate that RGB+depth methods are advantageous when maximising recall is the primary objective, whereas RGB-only methods such as MAEM are more practical when sensor deployment is constrained, and strict localisation precision is prioritised.

The core insight behind MAEM is that dense body surface geometry provides a substantially richer signal for cross-view association than either sparse keypoints or appearance features. When two players adopt similar poses, sparse keypoint configurations become nearly indistinguishable in epipolar space, whereas the dense mesh captures torso contours, limb thickness, and subtle postural variations that remain discriminative even under similar skeletal configurations. The ablation study quantitatively confirms this advantage. Furthermore, the two-stage filtering design is not merely a computational optimisation but a structural necessity. The reprojection filter effectively rejects spatially distant candidates but cannot distinguish between nearby players, whereas the mesh epipolar filter provides fine-grained discrimination but remains susceptible to degenerate epipolar configurations without prior spatial screening. Neither filter alone matches the full pipeline, confirming that their combination addresses a broader range of matching ambiguities.

The training-free nature of MAEM also proves to be a practical advantage. The framework generalises across two fundamentally different camera setups—an indoor gym with elevated fisheye cameras and an outdoor court with person-height cameras, without parameter retraining beyond threshold tuning. This contrasts with learning-based methods, which require domain-specific 3D annotations and are sensitive to changes in camera configuration.

Despite its strong performance, MAEM has several limitations that suggest directions for future work. First, its accuracy is inherently bounded by the monocular mesh recovery model: when SAM 3D Body fails to detect a player or produces inaccurate reconstructions due to severe occlusion or motion blur, downstream stages cannot recover. This contributes to the lower Recall on Human-M3 (83.7\%) compared to SportCenter EPFL (99.6\%). Second, the dense mesh epipolar filter introduces substantial computational overhead by requiring point-to-line distance computation for $N_v$ vertices per candidate pair, resulting in per-frame matching times exceeding one second on a single CPU thread. Although this operation is inherently parallelisable and could be accelerated via GPU batching, runtime efficiency remains a practical limitation of the current implementation. Reducing vertex count through intelligent subsampling or learned vertex selection therefore represents a promising optimisation direction. Finally, MAEM currently processes each frame independently. Incorporating temporal tracking could improve robustness against transient matching failures, enable consistent identity assignment across frames, and support downstream applications such as tactical analysis and player statistics.

\section{Conclusion}

Multi-view multi-person 3D pose estimation in team sports remains challenging for cross-view association. 
We proposed MAEM, a framework that addresses cross-view 
association by substituting these fragile signals with projected dense body 
mesh vertices as epipolar verification cues. The two-stage pipeline combines 
a lightweight bounding box reprojection filter with a dense mesh epipolar 
filter, followed by Hungarian matching and Union-Find clustering, enabling 
robust appearance-agnostic person association without any target-domain network training.
Experiments show that MAEM improves over training-free association baselines on two basketball benchmarks and achieves competitive RGB-only performance on Human-M3-Basketball.
On Sportcenter EPFL, MAEM achieves 99.6\% Recall and an MPJPE of 
59.8~mm. On Human-M3-Basketball, MAEM achieves 74.0~mm MPJPE and outperforms the evaluated RGB-only association baselines.
Despite these results, MAEM has notable limitations. Its accuracy is bounded 
by the monocular mesh recovery front-end, which may fail under severe 
occlusion or motion blur.
Beyond team sports, the core principle of MAEM, namely using dense geometric 
surface representations as cross-view association cues, may represent a 
promising direction for multi-camera scenarios where appearance and sparse 
keypoints are unreliable, though empirical validation in such settings 
remains an open question.

\section{Data Availability}

Both datasets used in this study are publicly available.
The Sportcenter EPFL Multiview dataset can be accessed at
\url{https://www.epfl.ch/labs/cvlab/data/sportcenter-dataset/}.
The Human-M3 dataset can be accessed at
\url{https://github.com/soullessrobot/Human-M3-Dataset}.

\section*{Acknowledgments}
    This work was financially supported by JST SPRING, Grant Number JPMJSP2125, JSPS Grant Number 23H03282, and JST PRESTO Grant Number JPMJPR20CA.
    The author L. Y. would like to take this opportunity to thank the ``THERS Make New Standards Program for the Next Generation Researchers''.

\section*{Declarations}
\subsection*{Conflict of Interest}
    The authors declare that they have no conflict of interest.



\bibliography{sn-bibliography}

@inproceedings{tu2020voxelpose,
  author    = {Tu, Hanyue and Wang, Chunyu and Zeng, Wenjun},
  title     = {VoxelPose: Towards Multi-Camera {3D} Human Pose Estimation in Wild Environment},
  booktitle = {Proceedings of the European Conference on Computer Vision (ECCV)},
  pages     = {197--212},
  year      = {2020}
}

@InProceedings{srivastav2024selfpose3d,
  author    = {Srivastav, Vinkle and Chen, Keqi and Padoy, Nicolas},
  title     = {SelfPose{3D}: Self-Supervised Multi-Person Multi-View {3D} Pose Estimation},
  booktitle = {Proceedings of the IEEE/CVF Conference on Computer Vision and Pattern Recognition (CVPR)},
  pages     = {2502--2512},
  year      = {2024}
}

@inproceedings{zhang2021direct,
  author    = {Wang, Tao and Zhang, Jianfeng and Cai, Yujun and Yan, Shuicheng and Feng, Jiashi},
  title     = {Direct Multi-View Multi-Person {3D} Pose Estimation},
  booktitle = {Advances in Neural Information Processing Systems},
  pages     = {13153--13164},
  year      = {2021}
}

@InProceedings{lin2021multi,
  author    = {Lin, Jiahao and Lee, Gim Hee},
  title     = {Multi-View Multi-Person {3D} Pose Estimation With Plane Sweep Stereo},
  booktitle = {Proceedings of the IEEE/CVF Conference on Computer Vision and Pattern Recognition (CVPR)},
  pages     = {11886--11895},
  year      = {2021}
}

@inproceedings{ye2022faster,
  author    = {Ye, Hang and Zhu, Wentao and Wang, Chunyu and Wu, Rujie and Wang, Yizhou},
  title     = {Faster VoxelPose: Real-Time {3D} Human Pose Estimation by Orthographic Projection},
  booktitle = {Proceedings of the European Conference on Computer Vision (ECCV)},
  pages     = {142--159},
  year      = {2022}
}

@inproceedings{huang2020end,
  author    = {Huang, Congzhentao and Jiang, Shuai and Li, Yang and Zhang, Ziyue and Traish, Jason and Deng, Chen and Ferguson, Sam and Da Xu, Richard Yi},
  title     = {End-to-End Dynamic Matching Network for Multi-View Multi-Person {3D} Pose Estimation},
  booktitle = {Proceedings of the European Conference on Computer Vision (ECCV)},
  pages     = {477--493},
  year      = {2020}
}

@inproceedings{liao2024multiple,
  author    = {Liao, Ziwei and Zhu, Jialiang and Wang, Chunyu and Hu, Han and Waslander, Steven L.},
  title     = {Multiple View Geometry Transformers for {3D} Human Pose Estimation},
  booktitle = {Proceedings of the IEEE/CVF Conference on Computer Vision and Pattern Recognition (CVPR)},
  pages     = {708--717},
  year      = {2024}
}

@inproceedings{qiu2023psvt,
  author    = {Qiu, Zhongwei and Yang, Qiansheng and Wang, Jian and Feng, Haocheng and Han, Junyu and Ding, Errui and Xu, Chang and Fu, Dongmei and Wang, Jingdong},
  title     = {{PSVT}: End-to-End Multi-Person {3D} Pose and Shape Estimation With Progressive Video Transformers},
  booktitle = {Proceedings of the IEEE/CVF Conference on Computer Vision and Pattern Recognition (CVPR)},
  pages     = {21254--21263},
  year      = {2023}
}

@inproceedings{he2023fastreid,
  author    = {He, Lingxiao and Liao, Xingyu and Liu, Wu and Liu, Xinchen and Cheng, Peng and Mei, Tao},
  title     = {{FastReID}: A PyTorch Toolbox for General Instance Re-Identification},
  booktitle = {Proceedings of the 31st ACM International Conference on Multimedia},
  pages     = {9664--9667},
  year      = {2023}
}

@inproceedings{he2021transreid,
  author    = {He, Shuting and Luo, Hao and Wang, Pichao and Wang, Fan and Li, Hao and Jiang, Wei},
  title     = {{TransReID}: Transformer-Based Object Re-Identification},
  booktitle = {Proceedings of the IEEE/CVF International Conference on Computer Vision (ICCV)},
  pages     = {15013--15022},
  year      = {2021}
}

@inproceedings{dong2019fast,
  author    = {Dong, Junting and Jiang, Wen and Huang, Qixing and Bao, Hujun and Zhou, Xiaowei},
  title     = {Fast and Robust Multi-Person {3D} Pose Estimation From Multiple Views},
  booktitle = {Proceedings of the IEEE/CVF Conference on Computer Vision and Pattern Recognition (CVPR)},
  pages     = {7792--7801},
  year      = {2019}
}

@inproceedings{zhang20204d,
  author    = {Zhang, Yuxiang and An, Liang and Yu, Tao and Li, Xiu and Li, Kun and Liu, Yebin},
  title     = {{4D} Association Graph for Realtime Multi-Person Motion Capture Using Multiple Video Cameras},
  booktitle = {Proceedings of the IEEE/CVF Conference on Computer Vision and Pattern Recognition (CVPR)},
  pages     = {1324--1333},
  year      = {2020}
}

@InProceedings{bridgeman2019multi,
  author    = {Bridgeman, Lewis and Volino, Marco and Guillemaut, Jean-Yves and Hilton, Adrian},
  title     = {Multi-Person {3D} Pose Estimation and Tracking in Sports},
  booktitle = {Proceedings of the IEEE/CVF Conference on Computer Vision and Pattern Recognition (CVPR) Workshops},
  year      = {2019}
}

@misc{sportcenter_multiview,
  author       = {{EPFL CVLAB}},
  title        = {{SportCenter Multi-View Human Pose Estimation Dataset}},
  howpublished = {\url{https://www.epfl.ch/labs/cvlab/data/sportcenter-dataset/}},
  year         = {2022},
  note         = {Accessed: April 19, 2026}
}

@inproceedings{dong2021shape,
  author    = {Dong, Zijian and Song, Jie and Chen, Xu and Guo, Chen and Hilliges, Otmar},
  title     = {Shape-Aware Multi-Person Pose Estimation From Multi-View Images},
  booktitle = {Proceedings of the IEEE/CVF International Conference on Computer Vision (ICCV)},
  pages     = {11158--11168},
  year      = {2021}
}

@inproceedings{hokari2025human,
  author    = {Hokari, Yamato and Hori, Ryosuke and Saito, Hideo},
  title     = {Human Mesh Reconstruction of Sports Players with Multiple Dynamic Cameras},
  booktitle = {Proceedings of the IEEE/CVF Conference on Computer Vision and Pattern Recognition (CVPR)},
  pages     = {6049--6059},
  year      = {2025}
}

@article{SMPL:2015,
  author    = {Loper, Matthew and Mahmood, Naureen and Romero, Javier and Pons-Moll, Gerard and Black, Michael J.},
  title     = {{SMPL}: A Skinned Multi-Person Linear Model},
  journal   = {ACM Trans. Graphics (Proc. SIGGRAPH Asia)},
  volume    = {34},
  number    = {6},
  pages     = {248:1--248:16},
  publisher = {ACM},
  year      = {2015}
}

@inproceedings{zhou2022quickpose,
  author    = {Zhou, Zhize and Shuai, Qing and Wang, Yize and Fang, Qi and Ji, Xiaopeng and Li, Fashuai and Bao, Hujun and Zhou, Xiaowei},
  title     = {QuickPose: Real-Time Multi-View Multi-Person Pose Estimation in Crowded Scenes},
  booktitle = {ACM SIGGRAPH 2022 Conference Proceedings},
  pages     = {1--9},
  year      = {2022}
}

@article{yang2026sam3dbody,
  author  = {Yang, Xitong and Kukreja, Devansh and Pinkus, Don and Sagar, Anushka and Fan, Taosha and Park, Jinhyung and Shin, Soyong and Cao, Jinkun and Liu, Jiawei and Ugrinovic, Nicolas and Feiszli, Matt and Malik, Jitendra and Dollar, Piotr and Kitani, Kris},
  title   = {{SAM 3D Body}: Robust Full-Body Human Mesh Recovery},
  journal = {arXiv preprint arXiv:2602.15989},
  year    = {2026}
}

@inproceedings{pavlakos2019expressive,
  author    = {Pavlakos, Georgios and Choutas, Vasileios and Ghorbani, Nima and Bolkart, Timo and Osman, Ahmed A. A. and Tzionas, Dimitrios and Black, Michael J.},
  title     = {Expressive Body Capture: {3D} Hands, Face, and Body From a Single Image},
  booktitle = {Proceedings of the IEEE/CVF Conference on Computer Vision and Pattern Recognition (CVPR)},
  pages     = {10975--10985},
  year      = {2019}
}

@article{MHR:2025,
  author        = {Aaron Ferguson and Ahmed A. A. Osman and Berta Bescos and Carsten Stoll and Chris Twigg and Christoph Lassner and David Otte and Eric Vignola and Fabian Prada and Federica Bogo and Igor Santesteban and Javier Romero and Jenna Zarate and Jeongseok Lee and Jinhyung Park and Jinlong Yang and John Doublestein and Kishore Venkateshan and Kris Kitani and Ladislav Kavan and Marco Dal Farra and Matthew Hu and Matthew Cioffi and Michael Fabris and Michael Ranieri and Mohammad Modarres and Petr Kadlecek and Rawal Khirodkar and Rinat Abdrashitov and Romain Pr{\'e}vost and Roman Rajbhandari and Ronald Mallet and Russell Pearsall and Sandy Kao and Sanjeev Kumar and Scott Parrish and Shoou-I Yu and Shunsuke Saito and Takaaki Shiratori and Te-Li Wang and Tony Tung and Yichen Xu and Yuan Dong and Yuhua Chen and Yuanlu Xu and Yuting Ye and Zhongshi Jiang},
  title         = {{MHR}: Momentum Human Rig},
  year          = {2025},
  journal={arXiv preprint arXiv:2511.15586}
}

@article{fischler1981random,
  author    = {Fischler, Martin A and Bolles, Robert C},
  title     = {Random Sample Consensus: A Paradigm for Model Fitting with Applications to Image Analysis and Automated Cartography},
  journal   = {Communications of the ACM},
  volume    = {24},
  number    = {6},
  pages     = {381--395},
  year      = {1981},
  publisher = {ACM New York, NY, USA}
}

@article{fan2023human,
  author  = {Fan, Bohao and Wang, Siqi and Zheng, Wenzhao and Feng, Jianjiang and Zhou, Jie},
  title   = {Human-{M3}: A Multi-View Multi-Modal Dataset for {3D} Human Pose Estimation in Outdoor Scenes},
  journal = {arXiv preprint arXiv:2308.00628},
  year    = {2023}
}

@misc{xrmocap,
  author       = {{XRMoCap Contributors}},
  title        = {OpenXRLab Multi-View Motion Capture Toolbox and Benchmark},
  howpublished = {\url{https://github.com/openxrlab/xrmocap}},
  year         = {2022}
}

@inproceedings{xu2022vitpose,
  author    = {Xu, Yufei and Zhang, Jing and Zhang, Qiming and Tao, Dacheng},
  title     = {Vi{TP}ose: Simple Vision Transformer Baselines for Human Pose Estimation},
  booktitle = {Advances in Neural Information Processing Systems},
  volume={35},
  pages={38571--38584},
  year      = {2022}
}

@article{kuhn1955hungarian,
  author    = {Kuhn, Harold W},
  title     = {The Hungarian Method for the Assignment Problem},
  journal   = {Naval Research Logistics Quarterly},
  volume    = {2},
  number    = {1-2},
  pages     = {83--97},
  year      = {1955},
  publisher = {Wiley Online Library}
}

@book{hartley2003multiple,
  author    = {Hartley, Richard and Zisserman, Andrew},
  title     = {Multiple View Geometry in Computer Vision},
  edition   = {2},
  address   = {Cambridge},
  publisher = {Cambridge University Press},
  year      = {2003}
}

@article{galler1964improved,
  author    = {Galler, Bernard A and Fisher, Michael J},
  title     = {An Improved Equivalence Algorithm},
  journal   = {Communications of the ACM},
  volume    = {7},
  number    = {5},
  pages     = {301--303},
  year      = {1964},
  publisher = {ACM New York, NY, USA}
}

@inproceedings{roy2022triangulation,
  author    = {Roy, Soumava Kumar and Citraro, Leonardo and Honari, Sina and Fua, Pascal},
  title     = {On Triangulation as a Form of Self-Supervision for {3D} Human Pose Estimation},
  booktitle = {2022 International Conference on {3D} Vision (3DV)},
  pages     = {1--10},
  year      = {2022}
}

@inproceedings{yeung2025athletepose3d,
  author    = {Yeung, Calvin and Suzuki, Tomohiro and Tanaka, Ryota and Yin, Zhuoer and Fujii, Keisuke},
  title     = {AthletePose{3D}: A Benchmark Dataset for {3D} Human Pose Estimation and Kinematic Validation in Athletic Movements},
  booktitle = {Proceedings of the IEEE/CVF Conference on Computer Vision and Pattern Recognition (CVPR) Workshops},
  pages     = {5991--6002},
  year      = {2025}
}

@inproceedings{belagiannis20143d,
  author    = {Belagiannis, Vasileios and Amin, Sikandar and Andriluka, Mykhaylo and Schiele, Bernt and Navab, Nassir and Ilic, Slobodan},
  title     = {{3D} Pictorial Structures for Multiple Human Pose Estimation},
  booktitle = {Proceedings of the IEEE Conference on Computer Vision and Pattern Recognition (CVPR)},
  pages     = {1669--1676},
  year      = {2014}
}

@inproceedings{Joo_2015_ICCV,
  author    = {Joo, Hanbyul and Liu, Hao and Tan, Lei and Gui, Lin and Nabbe, Bart and Matthews, Iain and Kanade, Takeo and Nobuhara, Shohei and Sheikh, Yaser},
  title     = {Panoptic Studio: A Massively Multiview System for Social Motion Capture},
  booktitle = {Proceedings of the IEEE International Conference on Computer Vision (ICCV)},
  pages     = {3334--3342},
  year      = {2015}
}

@inproceedings{jiang2024worldpose,
  author    = {Jiang, Tianjian and Billingham, Johsan and M{\"u}ksch, Sebastian and Zarate, Juan and Evans, Nicolas and Oswald, Martin R. and Polleyfeys, Marc and Hilliges, Otmar and Kaufmann, Manuel and Song, Jie},
  title     = {WorldPose: A World Cup Dataset for Global {3D} Human Pose Estimation},
  booktitle = {Proceedings of the European Conference on Computer Vision (ECCV)},
  pages     = {343--362},
  year      = {2025}
}

@inproceedings{wandt2021canonpose,
  author    = {Wandt, Bastian and Rudolph, Marco and Zell, Petrissa and Rhodin, Helge and Rosenhahn, Bodo},
  title     = {CanonPose: Self-Supervised Monocular {3D} Human Pose Estimation in the Wild},
  booktitle = {Proceedings of the IEEE/CVF Conference on Computer Vision and Pattern Recognition (CVPR)},
  pages     = {13294--13304},
  year      = {2021}
}

@inproceedings{bartol2022generalizable,
  author    = {Bartol, Kristijan and Bojani\'c, David and Petkovi\'c, Tomislav and Pribani\'c, Tomislav},
  title     = {Generalizable Human Pose Triangulation},
  booktitle = {Proceedings of the IEEE/CVF Conference on Computer Vision and Pattern Recognition (CVPR)},
  pages     = {11028--11037},
  year      = {2022}
}

@inproceedings{lu2024avatarpose,
  author    = {Lu, Feichi and Dong, Zijian and Song, Jie and Hilliges, Otmar},
  title     = {AvatarPose: Avatar-Guided {3D} Pose Estimation of Close Human Interaction from Sparse Multi-View Videos},
  booktitle = {Proceedings of the European Conference on Computer Vision (ECCV)},
  pages     = {215--233},
  year      = {2025}
}

@inproceedings{yin2025easyret3d,
  author    = {Yin, Junjie Oscar and Li, Ting and Wang, Jiahao and Zhang, Yi and Yuille, Alan},
  title     = {EasyRet{3D}: Uncalibrated Multi-View Multi-Human {3D} Reconstruction and Tracking},
  booktitle = {Proceedings of the IEEE/CVF Winter Conference on Applications of Computer Vision (WACV)},
  pages     = {3128--3137},
  year      = {2025}
}

@inproceedings{ingwersen2023sportspose,
  author    = {Ingwersen, Christian Keilstrup and Mikkelstrup, Christian M{\o}ller and Jensen, Janus N{\o}rtoft and Hannemose, Morten Rieger and Dahl, Anders Bjorholm},
  title     = {SportsPose -- A Dynamic {3D} Sports Pose Dataset},
  booktitle = {Proceedings of the IEEE/CVF Conference on Computer Vision and Pattern Recognition (CVPR) Workshops},
  pages     = {5219--5228},
  year      = {2023}
}

@inproceedings{suzuki2025athleticspose,
  author    = {Suzuki, Tomohiro and Tanaka, Ryota and Yeung, Calvin and Fujii, Keisuke},
  title     = {AthleticsPose: Authentic Sports Motion Dataset on Athletic Field and Evaluation of Monocular {3D} Pose Estimation Ability},
  booktitle = {Proceedings of the 8th International ACM Workshop on Multimedia Content Analysis in Sports},
  pages     = {8--17},
  year      = {2025}
}

@inproceedings{yeung2024autosoccerpose,
  author    = {Yeung, Calvin and Ide, Kenjiro and Fujii, Keisuke},
  title     = {AutoSoccerPose: Automated {3D} Posture Analysis of Soccer Shot Movements},
  booktitle = {Proceedings of the IEEE/CVF Conference on Computer Vision and Pattern Recognition (CVPR) Workshops},
  pages     = {3214--3224},
  year      = {2024}
}

@article{yin2024enhanced,
  author  = {Yin, Li and Yeung, Calvin and Hu, Qingrui and Ichikawa, Jun and Azechi, Hirotsugu and Takahashi, Susumu and Fujii, Keisuke},
  title   = {Enhanced Multi-Object Tracking Using Pose-Based Virtual Markers in {3x3} Basketball},
  journal = {Sports Engineering},
  volume  = {29},
  number  = {1},
  pages   = {12},
  year    = {2026}
}

@inproceedings{yamada2025trackid3x3,
  author    = {Yamada, Kazuhiro and Yin, Li and Hu, Qingrui and Ding, Ning and Iwashita, Shunsuke and Ichikawa, Jun and Kotani, Kiwamu and Yeung, Calvin and Fujii, Keisuke},
  title     = {{TrackID3x3}: A Dataset and Algorithm for Multi-Player Tracking with Identification and Pose Estimation in {3x3} Basketball Full-Court Videos},
  booktitle = {Proceedings of the 8th International ACM Workshop on Multimedia Content Analysis in Sports},
  pages     = {163--173},
  year      = {2025}
}

\newpage
\section{Supplementary Material}
Table~\ref{tab:S1} presents the results on the Human-M3 dataset.

\begin{table}[htbp]
\centering
\renewcommand{\thetable}{S\arabic{table}}
\setcounter{table}{0}
\caption{Comparison with state-of-the-art 3D human pose estimation algorithms. Metrics include MPJPE (mm), Recall (500 mm) and Average Precision (AP). Best values are shown in \textbf{bold}. (PCD: point cloud input.) Reproduced from the original paper.}
\label{tab:S1}
\small
\setlength{\tabcolsep}{4pt}
\begin{tabular}{lcccccccc}
\hline
Algorithm &Input Modality & MPJPE & Recall & AP$_{75}$ & AP$_{100}$ & AP$_{125}$ & AP$_{150}$ \\
\hline
VoxelPose~\cite{tu2020voxelpose}      & RGB      & 108.0 & 90.8 & 23.0 & 44.9 & 59.7 & 68.5 \\
PlaneSweepPose~\cite{lin2021multi} & RGB      &  99.0 & 39.8 &  4.7 &  8.4 & 10.9 & 12.1 \\
MVP~\cite{zhang2021direct}            & RGB      & 140.0 & 89.3 &  7.4 & 22.3 & 37.7 & 50.1 \\
MMVP~\cite{fan2023human}          & RGB+PCD & \textbf{79.0} & \textbf{98.3} & \textbf{41.9} & \textbf{69.9} & \textbf{81.5} & \textbf{87.7} \\
\hline
\end{tabular}
\end{table}

Figure~\ref{fig:s1} provides a detailed flowchart of the MAEM pipeline, complementing the overview in Figure 1 with explicit decision branches and threshold conditions at each processing step.

\begin{figure*}[t]
\centering
\renewcommand{\thefigure}{S\arabic{figure}}
\setcounter{figure}{0}
\includegraphics[width=\textwidth]{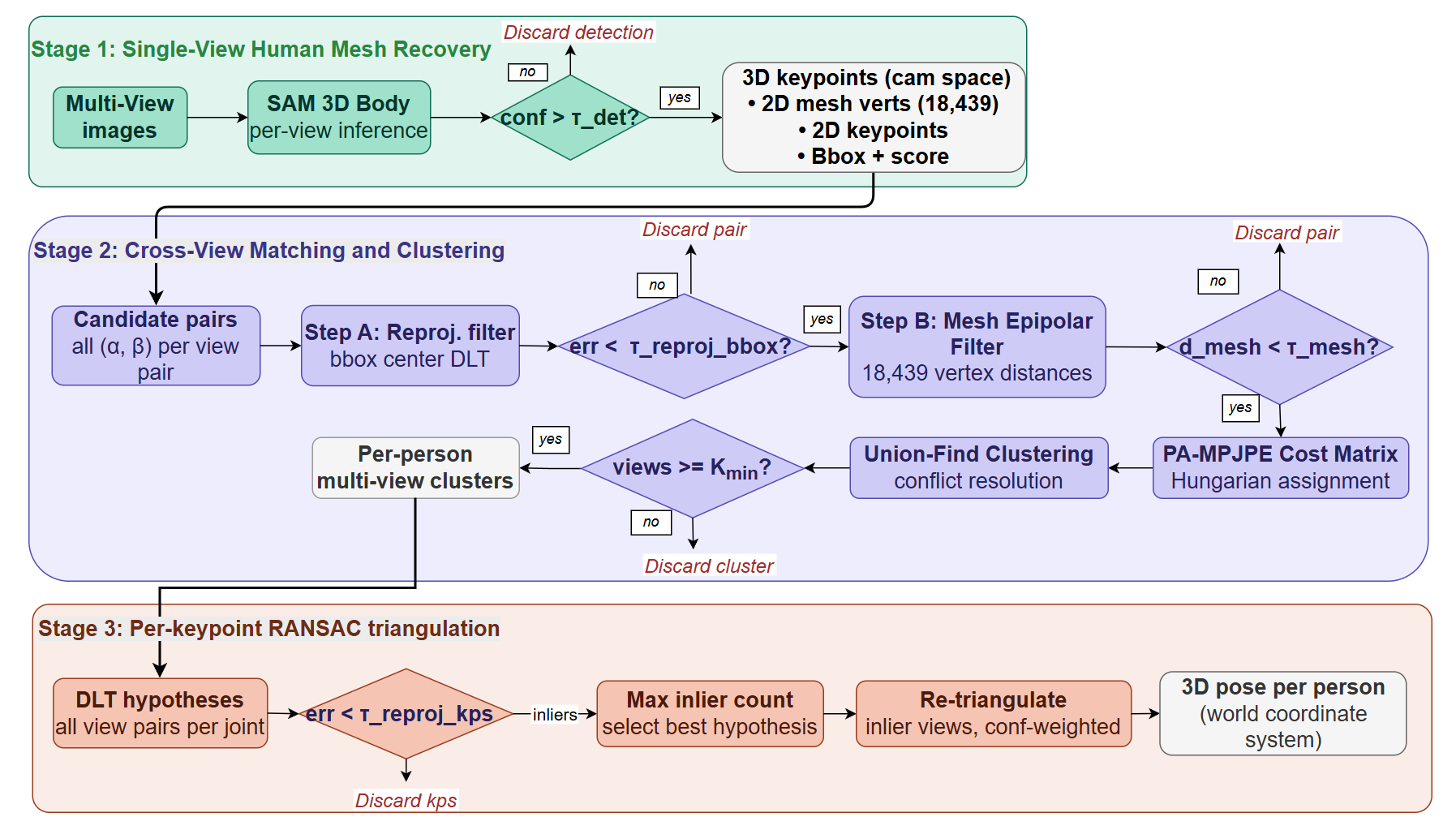}
\caption{Detailed flowchart of the MAEM pipeline, where each processing step is described textually. Given multi-view images, Stage 1 recovers per-person 3D meshes and keypoints via SAM 3D Body. Stage 2 associates detections across views using a two-stage epipolar filter and Hungarian matching. Stage 3 reconstructs the final 3D pose via RANSAC triangulation.}
\label{fig:s1}
\end{figure*}

\begin{figure}[htbp]
    \centering
    \renewcommand{\thefigure}{S\arabic{figure}}
    \setlength{\tabcolsep}{2pt}
    \begin{tabular}{cccccccc}
    \includegraphics[width=0.11\linewidth]{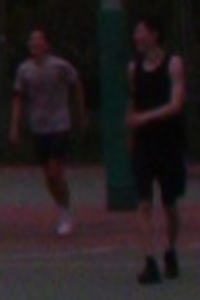} &
    \includegraphics[width=0.11\linewidth]{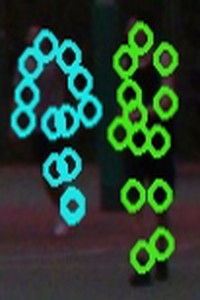} &
    \includegraphics[width=0.11\linewidth]{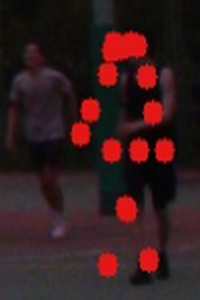} &
    \includegraphics[width=0.11\linewidth]{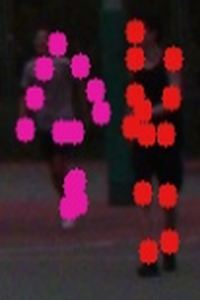} &
    \includegraphics[width=0.11\linewidth]{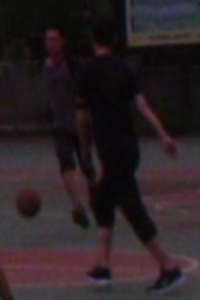} &
    \includegraphics[width=0.11\linewidth]{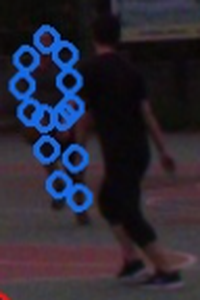} &
    \includegraphics[width=0.11\linewidth]{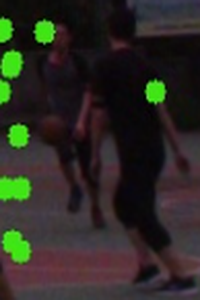} &
    \includegraphics[width=0.11\linewidth]{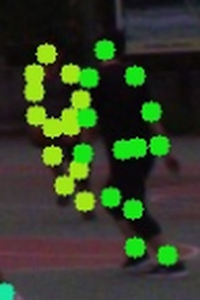} \\[2pt]
    \includegraphics[width=0.11\linewidth]{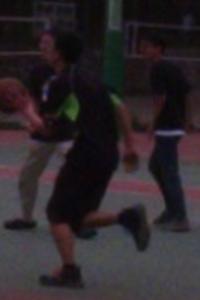} &
    \includegraphics[width=0.11\linewidth]{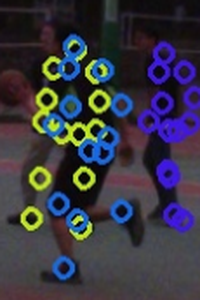} &
    \includegraphics[width=0.11\linewidth]{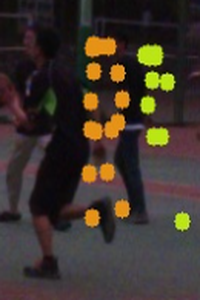} &
    \includegraphics[width=0.11\linewidth]{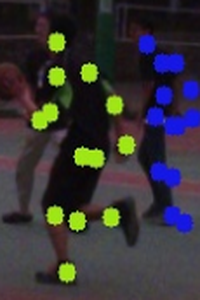} &
    \includegraphics[width=0.11\linewidth]{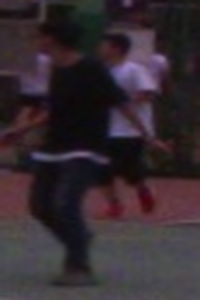} &
    \includegraphics[width=0.11\linewidth]{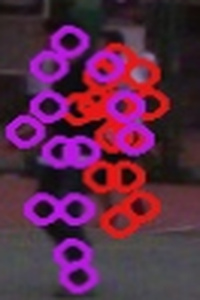} &
    \includegraphics[width=0.11\linewidth]{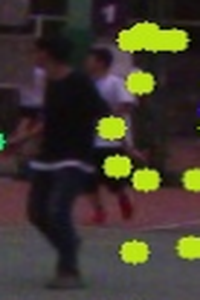} &
    \includegraphics[width=0.11\linewidth]{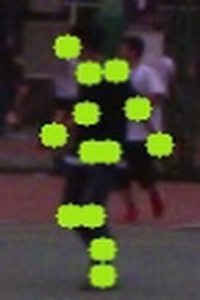} \\[2pt]
    \includegraphics[width=0.11\linewidth]{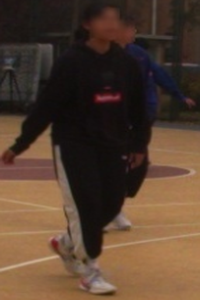} &
    \includegraphics[width=0.11\linewidth]{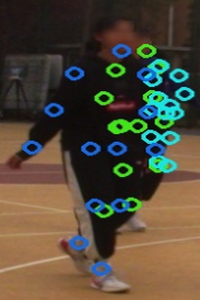} &
    \includegraphics[width=0.11\linewidth]{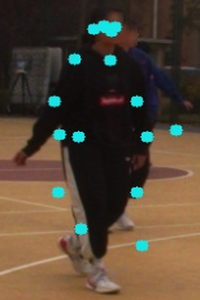} &
    \includegraphics[width=0.11\linewidth]{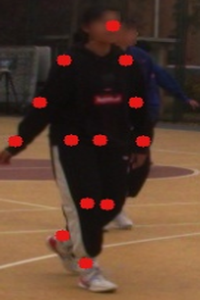} &
    \includegraphics[width=0.11\linewidth]{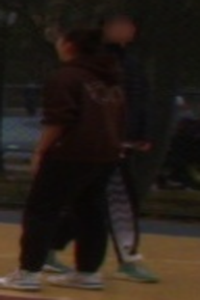} &
    \includegraphics[width=0.11\linewidth]{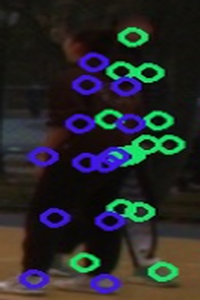} &
    \includegraphics[width=0.11\linewidth]{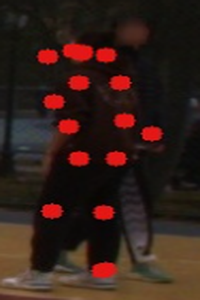} &
    \includegraphics[width=0.11\linewidth]{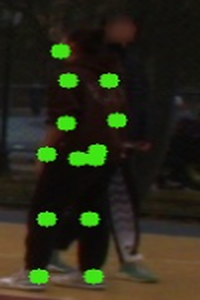} \\[2pt]
    \includegraphics[width=0.11\linewidth]{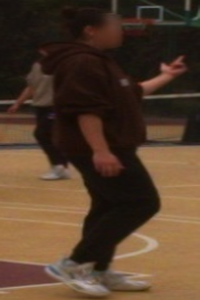} &
    \includegraphics[width=0.11\linewidth]{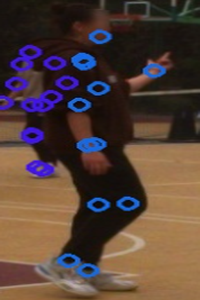} &
    \includegraphics[width=0.11\linewidth]{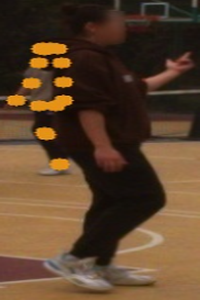} &
    \includegraphics[width=0.11\linewidth]{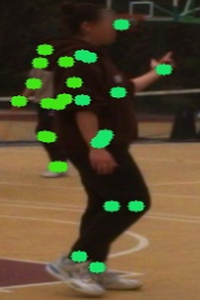} &
    \includegraphics[width=0.11\linewidth]{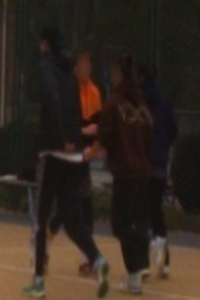} &
    \includegraphics[width=0.11\linewidth]{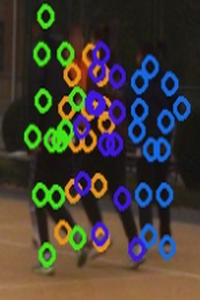} &
    \includegraphics[width=0.11\linewidth]{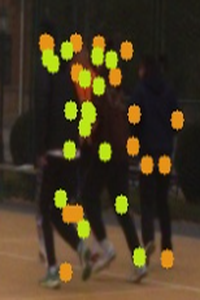} &
    \includegraphics[width=0.11\linewidth]{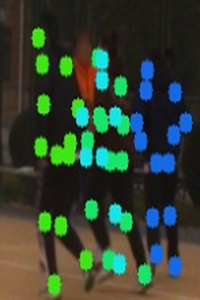} \\[2pt]
    Input & GT & MVPose & MAEM &Input & GT & MVPose & MAEM \\
\end{tabular}
    \caption{Additional qualitative results on the two evaluated datasets. Each row shows, from left to right, the input image, ground truth, MVPose prediction, and MAEM prediction. Rows 1–2 correspond to Human-M3 Basketball1, and rows 3–4 correspond to Human-M3 Basketball2.}
    \label{fig:s2} 
\end{figure}

Figure~\ref{fig:s2} presents additional qualitative results of MAEM 3D pose reprojections.

\end{document}